\documentclass{ieeeaccess}
\usepackage{cite}
\usepackage{amsmath,amssymb,amsfonts}
\usepackage{algorithm}
\usepackage{algorithmic}
\usepackage{graphicx}
\usepackage{subfigure}
\usepackage{textcomp}
\usepackage{adjustbox}

\usepackage{url}

\newcommand{\din}{\mathcal D}

\def\BibTeX{{\rm B\kern-.05em{\sc i\kern-.025em b}\kern-.08em
    T\kern-.1667em\lower.7ex\hbox{E}\kern-.125emX}}
\begin{document}
\history{This paper has been accepted by IEEE Access.}
\doi{10.1109/ACCESS.2017.DOI}

\title{RAILS: A Robust Adversarial Immune-inspired Learning 
System}
\author{\uppercase{Ren Wang}\authorrefmark{1}, 
\uppercase{Tianqi Chen\authorrefmark{1}, Stephen Lindsly}.\authorrefmark{1}, \uppercase{Cooper Stansbury\authorrefmark{1}, Alnawaz Rehemtulla}.\authorrefmark{1}, \uppercase{Indika Rajapakse}.\authorrefmark{1} and \uppercase{Alfred Hero}\authorrefmark{1} \IEEEmembership{Life Fellow, IEEE}}
\address[1]{University of Michigan, Ann Arbor, MI 48109 USA (e-mail: renwang@umich.edu, tqch@umich.edu, lindsly@umich.edu, cstansbu@umich.edu, alnawaz@umich.edu, indikar@umich.edu, hero@eecs.umich.edu)}
\tfootnote{This work was partially supported by the United States Dept of Defense, Defense Advanced Research Projects Agency (DARPA) under grant HR00112020011.} 

\markboth
{Author \headeretal: Preparation of Papers for IEEE TRANSACTIONS and JOURNALS}
{Author \headeretal: Preparation of Papers for IEEE TRANSACTIONS and JOURNALS}

\corresp{Corresponding authors: Alfred Hero (e-mail: hero@eecs.umich.edu); Indika Rajapakse (e-mail: indikar@umich.edu).}

\begin{abstract}
Adversarial attacks against deep neural networks (DNNs) are continuously evolving, requiring increasingly powerful defense strategies. We develop a novel adversarial defense framework inspired by the adaptive immune system: the Robust Adversarial Immune-inspired Learning System (RAILS). Initializing a population of exemplars that is balanced across classes, RAILS starts from a uniform label distribution that encourages diversity and uses an evolutionary optimization process to adaptively adjust the predictive label distribution in a manner that emulates the way the natural immune system recognizes novel pathogens. RAILS' evolutionary optimization process explicitly captures the tradeoff between robustness (diversity) and accuracy (specificity) of the network, and represents a new immune-inspired perspective on adversarial learning. The benefits of RAILS are empirically demonstrated under eight types of adversarial attacks on a DNN adversarial image classifier for several benchmark datasets, including: MNIST; SVHN; CIFAR-10; and CIFAR-10. We find that PGD is the most damaging attack strategy and that for this attack RAILS is significantly more robust than other methods, achieving improvements in adversarial robustness by $\geq 5.62\%, 12.5\%$, $10.32\%$, and $8.39\%$, on these respective datasets, without appreciable loss of classification accuracy. Codes for the results in this paper are available at \url{https://github.com/wangren09/RAILS}.
\end{abstract}

\begin{keywords}
Bio-inspired deep learning, adversarial robustness, deep network defense strategies.
\end{keywords}

\titlepgskip=-15pt

\maketitle

\section{Introduction}
\label{sec: intro}
\PARstart{T}{he} state of the art in supervised deep learning has dramatically improved over the past decade \cite{lecun2015deep}. Deep learning techniques have led to significant advances in applications such as: face recognition \cite{mehdipour2016comprehensive}; object detection \cite{zhao2019object}; and natural language processing \cite{young2018recent}. Despite these successes, deep learning techniques are not resilient to evasion attacks (a.k.a. adversarial attacks) on test inputs and poisoning attacks on training data \cite{goodfellow2014explaining,szegedy2013intriguing,GDG17}. The adversarial vulnerability of deep neural networks (DNN) have restricted their application, motivating researchers to develop effective defense methods. The focus of this paper is to develop a novel deep defense framework inspired by the mammalian immune system.

Current adversarial
defense strategies
can be divided into four classes: (1) detection of adversarial samples  \cite{metzen2017detecting,feinman2017detecting,grosse2017statistical}; (2) Robust training \cite{madry17,zhang2019theoretically,cohen2019certified,zhang2021robust}; (3) data denoising and reconstruction \cite{mustafa2019image,sun2019adversarial}; and (4) deep adversarial learning architectures \cite{papernot2018deep,samangouei2018defense}. The first class of methods defends a DNN using simple outlier 
detection models for detecting adversarial examples. However, it has been shown that such adversarial detection methods can be easily defeated \cite{carlini2017adversarial}. Robust training aims to harden the model to deactivate attacks such as evasion attacks. Known robust training methods are often tailored to a certain level of attack strength in the context of $\ell_p$-perturbation. Moreover, the trade-off between accuracy and robustness presents design challenges \cite{zhang2019theoretically}. The data denoising and reconstruction class of methods is driven by an intuitive 
idea that adversarial examples can be mapped to the manifold of clean examples through data reconstruction by denoising. However, while denoising can reduce adversarial perturbations it can also distort the inputs \cite{aldahdooh2021adversarial}, providing new opportunities for an attacker to exploit the defense mechanism \cite{niu2020limitations}. Deep adversarial learning architectures directly design the defense into the layers of the neural network, e.g., by robustifying them with $k$-NN's \cite{papernot2018deep} or modifying a generative adversarial network (GAN) \cite{samangouei2018defense}. Despite these advances, current methods still have difficulty providing an acceptable level of robustness to novel attacks \cite{athalye2018obfuscated}.

\begin{figure}[h]
\centerline{\includegraphics[trim=40 0 0 0,clip,width=0.95\linewidth]{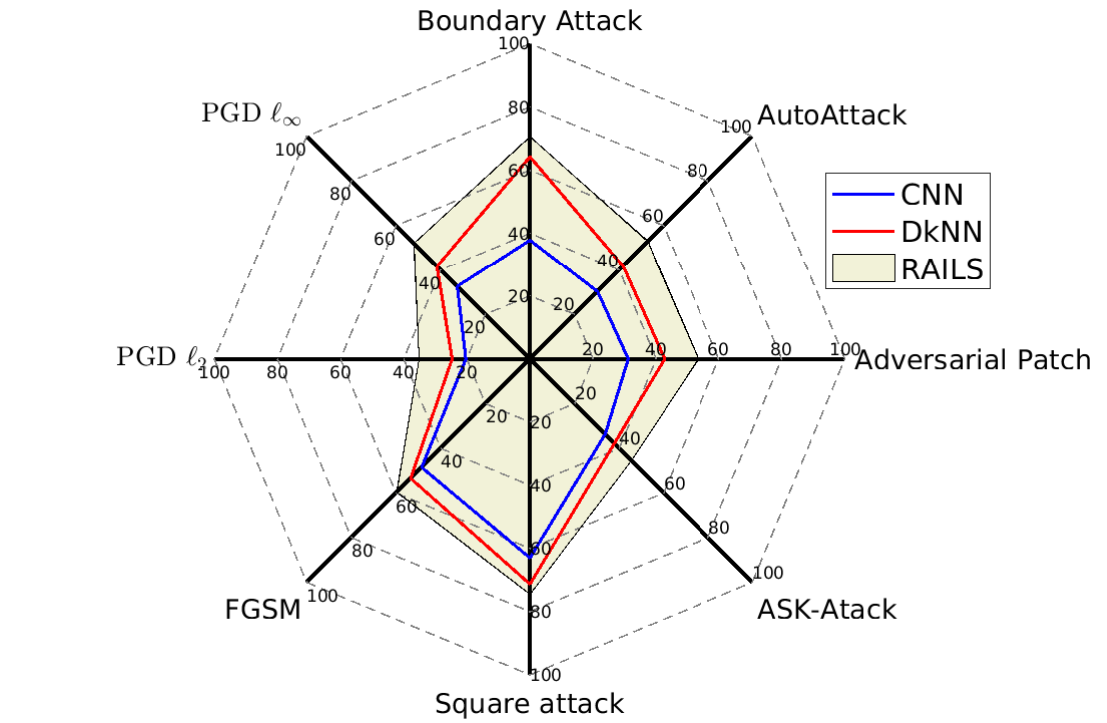}}
\caption{\textbf{RAILS launches the best defense against different types of attacks.} Radar plot showing that RAILS has higher robust accuracy than the adversarially trained CNN \cite{madry17} and Deep k-Nearest Neighbor (DkNN) \cite{papernot2018deep} in defending against eight types of attacks: $\ell_\infty$-PGD attack/ $\ell_2$-PGD attack \cite{madry17}, Fast Gradient Sign Method (FGSM) \cite{goodfellow2014explaining}, Square Attack \cite{andriushchenko2020square}, Adversarial Patch \cite{brown2017adversarial}, AutoAttack \cite{croce2020reliable}, Boundary Attack \cite{brendel2017decision}, and a (customized) ASK-Attack \cite{wang2021ask}. The benign accuracy for CNN, DkNN, and RAILS are $87.26\%$, $86.63\%$, and $82\%$. We refer readers to Section~\ref{sec: experiments} for more results.}
\label{fig: overview_result}
\end{figure}

To design an effective defense, it is natural to consider a learning strategy that emulates mechanisms of the naturally robust biological immune system. The power of  artificial immune system (AIS) models have been established in many other applications \cite{jamali2017dawa,scaranti2020artificial,tarao2016toward}. While AIS approaches to enhancing DNN adversarial robustness have been previously developed  \cite{gupta2021using,rademaker2019attack}, they have been restricted to simple emulations of the innate immune system. In this paper, we propose a new framework, Robust Adversarial Immune-inspired Learning System (RAILS), that can effectively defend deep learning architectures against aggressive attacks based on a refined biology model of the adaptive immune system. Built upon a class-wise k-Nearest Neighbor (kNN) structure, given a test sample RAILS finds an initial small population of proximal samples, balanced across different classes, with uniform label distribution. RAILS then promotes the specificity of the label distribution towards the ground truth label through an evolutionary optimization. RAILS can efficiently correct adversarial label flipping by balancing label diversity against specificity. While RAILS can be applied to defending against many different types of attacks, in this paper we restrict attention to evasion attacks on the input. Figure~\ref{fig: overview_result} shows that RAILS outperforms existing methods on various types of evasion attacks.

Compared to existing defense methods, we make the \textbf{following contributions}:

$\bullet$ RAILS achieves better adversarial robustness by assigning a uniform label distribution to each input and evolving it to a distribution that is concentrated about the input's true label class. (see Section~\ref{sec: RAILS} and Table~\ref{tab2: acc_overall})

$\bullet$ RAILS incorporates a life-long robustifying process by adding synthetic ``virtual data'' to the training data. (see Section~\ref{sec: RAILS} and Table~\ref{tab4: acc_cnn_hard})

$\bullet$ RAILS evolves the distribution via mutation and cross-over mechanisms and is not restricted to $\ell_p$ or any other specific type of attack. (see Section~\ref{sec: alg_detail} and Figure~\ref{fig: overview_result})

$\bullet$ We demonstrate that RAILS improves robustness of existing methods for different types of attacks (Figure~\ref{fig: overview_result}). In particular, RAILS improves robustness against the highly damaging PGD attack by $\geq 5.62\%/12.5\%/10.32\%/8.39\%$ for the  MNIST, SVHN, CIFAR-10 and CIFAR-100 datasets (Table~\ref{tab2: acc_overall}). Furthermore, we show that the RAILS implementation of life-long learning with training data augmentation yields  a $2.3\%$ robustness improvement 
with only $5\%$ augmentation of the training data  (Table~\ref{tab4: acc_cnn_hard}).

\begin{figure*}[h]
\centerline{\includegraphics[width=.85\textwidth]{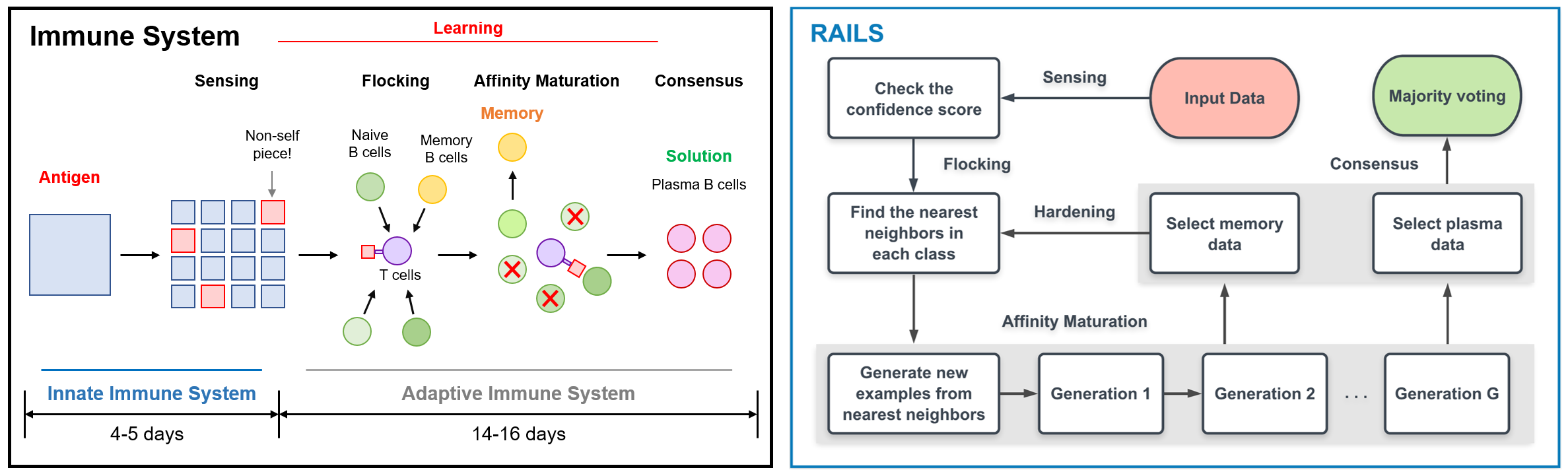}}
\caption{\textbf{Simplified immune system (left) and RAILS computational workflow (right):} Both systems are composed of a four-step process, which includes initial detection (sensing), recruiting candidates for diversity (flocking), enlarging population size and promoting specificity (affinity maturation) \cite{de2015dynamics}, and obtaining the final solution (consensus).}
\label{fig: bio_comp}
\end{figure*}

$\bullet$ RAILS is the first adversarial defense framework to be based on the biology of the adaptive immune system. In particular: (a) RAILS computationally emulates the principal mechanisms of the immune response (Figure~\ref{fig: bio_comp}); and (b) our computational and biological experiments demonstrate the fidelity of the emulation -  the learning patterns of RAILS and the immune system are closely aligned (Figure~\ref{fig: curves_bio_rails}).

\subsection{Related Work}
After it was established that DNNs were vulnerable to evasion attacks \cite{szegedy2013intriguing}, different types of defense mechanisms have been proposed. 
An intuitive idea is to eliminate the adversarial examples through outlier detection, including training an additional discrimination sub-network \cite{metzen2017detecting,grosse2017statistical} and using kernel density estimation \cite{feinman2017detecting}. The above approaches rely on the fundamental assumption that the distributions of benign and adversarial examples are easily distinguishable, an assumption that has been challenged in \cite{carlini2017adversarial}.

In addition to adversarial attack detection, other methods have been proposed that focus on robustifying the deep architecture during the learning phase \cite{madry17,cohen2019certified,zhang2019theoretically}. One recent approach combines training with perturbed inputs and  hierarchical feature alignment between the adversarial and clean domains to robustify the feature learning process \cite{zhang2021robust}. Though such defenses are effective against adversarial examples with moderate levels of $\ell_p$ attack strength, they have limited power to defend against stronger attacks, and there is often a sacrifice in overall classification accuracy. In contrast, RAILS is developed to defend against diverse powerful attacks with less sacrifice in accuracy, and can improve any model's robustness, including robust models.

A different family of defenses models adversarial inputs as deviating from the manifold of clean data. This motivates the use of projection methods for denoising the inputs, where the inputs are mapped to the manifold \cite{mustafa2019image,sun2019adversarial}. Examples include mapping adversarial examples to a high-resolution manifold using wavelet denoising and super resolution techniques \cite{mustafa2019image} and to a low dimensional quasi-natural space with a sparse transformation layer \cite{sun2019adversarial}. Despite the simple and clear motivation, denoising methods have their own
limitations. For example, they can introduce distortions into clean inputs \cite{aldahdooh2021adversarial} and have shown to fail to be robust to many adversarial attacks \cite{niu2020limitations,choi2019evaluating}. Instead of attempting to modify inputs, RAILS evolves a statistical population of clones of the input, resulting  in enhancing  resilience to attacks.

Another approach is to incorporate different architectures to robustify deep classifiers \cite{papernot2018deep,samangouei2018defense}. An example is the deep k-Nearest Neighbor (DkNN) classifier \cite{papernot2018deep} that robustifies against instance perturbations by applying kNN's to features extracted from each layer. However, a single kNN classifier applied on the whole dataset is easily to be fooled by strong attacks (Figure~\ref{fig: vis_compare_rails}). Conversely, RAILS incorporates an evolutionary \textit{diversity} to \textit{specificity} defense mechanism which can provide additional robustness to existing DNNs.

Network defense mechanisms inspired by the natural immune system have been proposed for other applications, different from the deep learning application considered in this paper. Among these,    artificial immune system (AIS) approaches  \cite{fernandes2017applications}  have been used to defend against wormhole attacks on mobile ad hoc networks \cite{jamali2017dawa}, flooding attacks on software-defined networks \cite{scaranti2020artificial}, and denial of service attacks on the internet \cite{tarao2016toward}. However, the closest point of tangency to our RAILS approach is recent work that borrows concepts from the innate immune system to detect adversarial examples in DNN's. The innate immune system, also known as the non-specific immune system, is nature's first line of defense that launches an immediate non-specific response to contain the pathogen using chemical cellular, and extracellular mechanisms to prevent pathogen mobility and spread. Mechanisms of innate immunity that have been emulated in machine learning include: negative selection algorithm approaches \cite{gupta2021using}; and cellular machanisms for early pathogen recognition \cite{rademaker2019attack}. Different from the innate immune system, the response of the adaptive immune system is longlasting, specific and sustained, using clonal expansion to produce B-cell and T-cell lymphocytes having antigen receptors specific to the pathogen. To the best of our knowledge, the proposed RAILS adversarial defense framework is the first to be based on the complex biology of the adaptive immune system.

Another line of research relevant to ours is adversarial transfer learning \cite{liu2019transferable,Shafahi2020Adversarially}, which aims to maintain robustness when there is covariate shift between training data and test data. We remark that covariate shift is naturally handled by the mutation mechanism in our adaptive immune system emulation of RAILS that adapts the defense to novel attacks.

\subsection{Learning Strategies of Immune System}

Systems robustness is a property that must be intentionally designed into the architecture, and one of the greatest examples of this is the mammalian adaptive immune system \cite{rajapakse2011emerging}. The adaptive immune system is incredibly complex and not something that we can hope to replicate at this time. However, we can simplify its robust learning process into these four steps: sensing, flocking, affinity maturation, and consensus (Figure \ref{fig: bio_comp} left) \cite{cucker2007emergent,rajapakse2017emergence}. The architecture of the adaptive immune system ensures a robust response to foreign antigens, splitting the work between active sensing and competitive growth to produce an effective antibody. \textit{Sensing} of a foreign attack leads to antigen-specific B-cells \textit{flocking} to some temporary structures for \textit{affinity maturation} \cite{de2015dynamics}. In the affinity maturation phase, a diverse initial set of antigen-specific B-cells divide to populate the temporary structures. Then the genetic identity of each B-cell is encoded by the shape and sequence of its protein, which can change from generation to generation. The degree to which the encoding of the B-cell matches the antigen is called the affinity. The B-cells with the highest affinity to the antigen are selected to divide and mutate, which leads to new B-cells with higher affinity to the antigen \cite{mesin2016germinal}. B-cells that reach \textit{consensus}, or achieve a threshold affinity against the foreign antigen, undergo terminal differentiation into plasma B-cells. Plasma B-cells represent the antigen-specific solutions. Memory B-cells are selected and stored to defend against similar attacks in the future.

\subsection{Notation and Preliminaries}
Given a mapping $f: \mathbb{R}^{d} \rightarrow \mathbb{R}^{d^{\prime}}$ and $\mathbf x_1, \mathbf x_2 \in \mathbb{R}^{d}$, we first define the affinity score $A(f;\mathbf x_1, \mathbf x_2)$ between $\mathbf x_1$ and $\mathbf x_2$. This affinity score can be defined in many ways, e.g, cosine similarity, inner product, or inverse $\ell_p$ distance, but here we use  $A(f;\mathbf x_1, \mathbf x_2) = -\|f(\mathbf x_1) - f(\mathbf x_2)\|_2$,  the negative Euclidean distance. In the context of DNN, $f$ denotes the feature mapping from input to feature representation, and $A$ measures the similarity between two inputs. Higher affinity score indicates higher similarity.

\section{RAILS: overview}\label{sec: RAILS}

\begin{figure}[h]
\centerline{\includegraphics[trim=0 0 0 0,clip,width=0.97\linewidth]{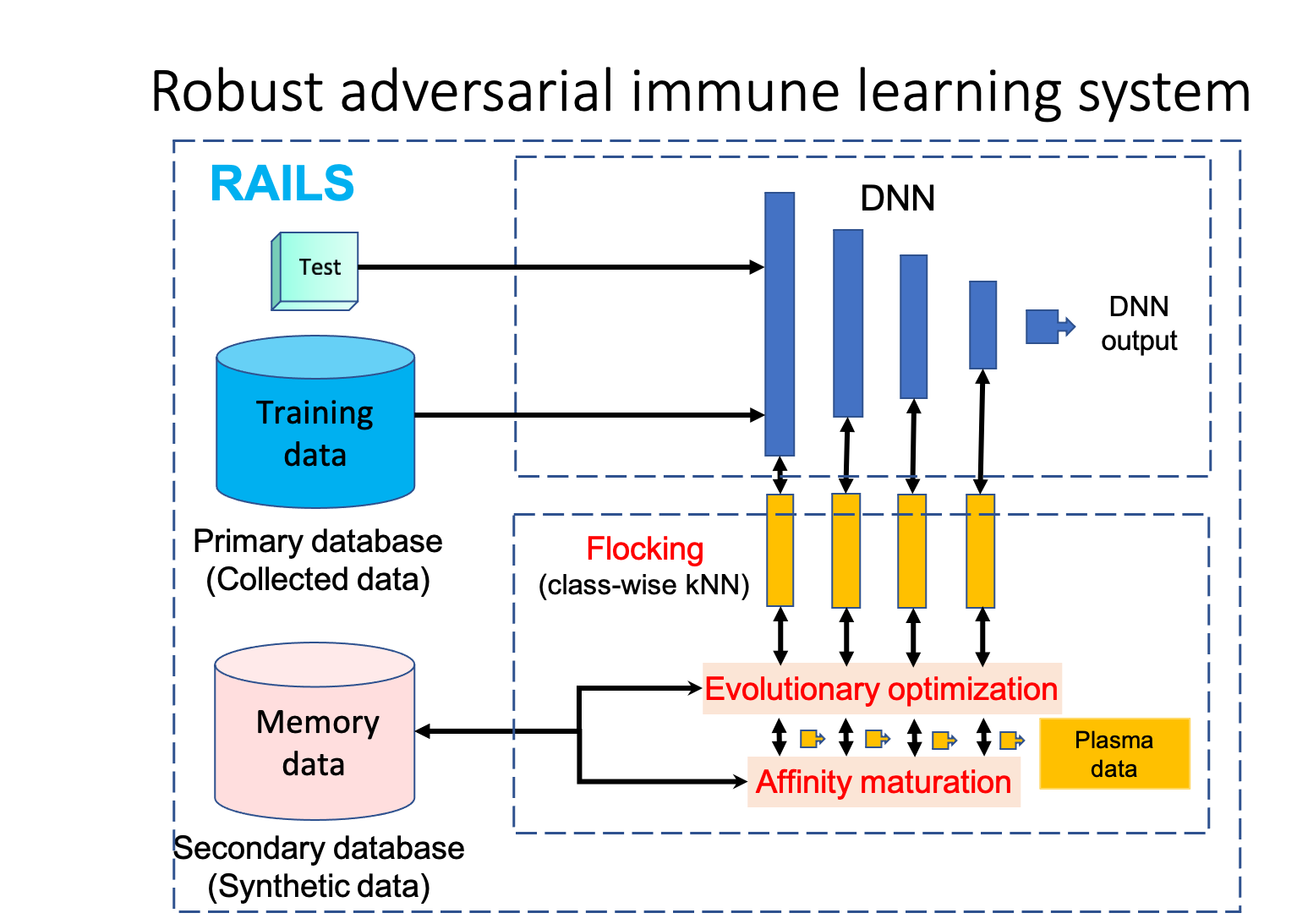}}
\caption{\textbf{The architecture of RAILS.} For each test input, a special type of data - plasma data - is generated by evolutionary optimization, that contributes to predicting the class of the input (test) sample. Another type of data - memory data - is generated and stored to help defend against similar attacks in the future. Plasma data and memory data are analogous to plasma B cells and memory B cells in the immune system.}
\label{fig: arch}
\end{figure}

In this section, we give an overview of RAILS, and provide a comparison to the natural immune system. The architecture of RAILS is illustrated in Figure~\ref{fig: arch}. For each selected hidden layer $l$, RAILS builds class-wise kNN architectures on training samples $\din_{tr}$. Then for each test input $\mathbf x$, a population of candidates is selected and goes through an evolutionary optimization process to obtain the optimal solution. In RAILS, two types of data are obtained after the evolutionary optimization: `plasma data' for optimal predictions of the present inputs, and `memory data' for the defense against future attacks. These two types of data correspond to plasma B cells and memory B cells in the biological system, and play important roles in \textit{static learning} and \textit{adaptive learning}, respectively.

\subsection{Defense with static learning}
 
\begin{figure}[h]
\centerline{\includegraphics[trim=20 0 0 0,clip,width=.49\textwidth]{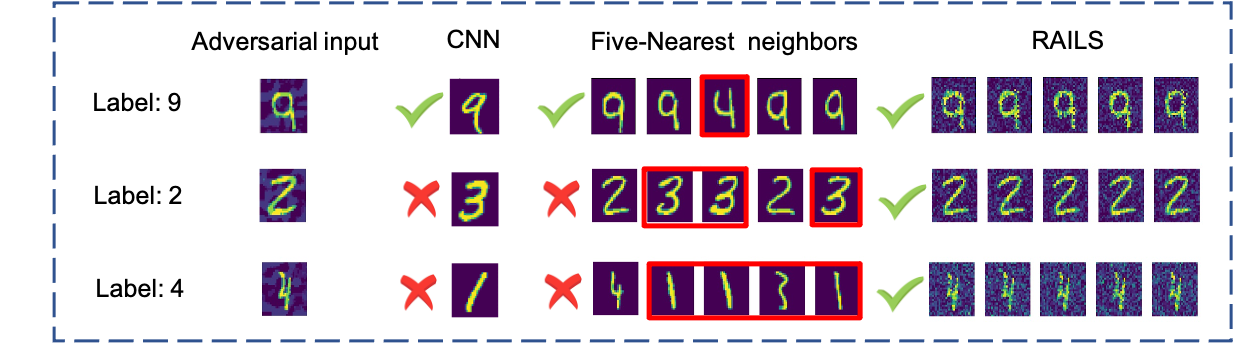}}
\caption{\textbf{Representation examples showing RAILS correcting (improving) the wrong (unconfident) predictions by CNN and kNN.} kNN predict all three examples with $\le 100\%$ confidence of the ground truth class, and CNN gets wrong predictions for images 2 and 4. RAILS provides correct predictions with high confidence scores.}
\label{fig: vis_compare_rails}
\end{figure}

Adversarial perturbations can severely affect deep classifiers, forcing the predictions to be dominated by adversarial classes rather than the ground truth. For example, a single kNN classifier is vulnerable to adversarial inputs, as shown in Figure~\ref{fig: vis_compare_rails}. The purpose of static learning is to address this issue, i.e., maintaining or increasing the prediction probability of the ground truth $p_l^{y_{\text{true}}(\mathbf x)}(\mathbf x)$ when the input $\mathbf x$ is manipulated by an adversary. The key components include \textbf{(i)} a label initialization via class-wise kNN search on $\din_{tr}$ that guarantees labels across different classes are uniformly distributed for each input; and \textbf{(ii)} an evolutionary data-label optimization that promotes label distribution specificity towards the input's true label class. Our hypothesis is that the covariate shift of the adversarial examples from the distribution of the ground truth class is small in the input space, and therefore new examples inherited from parents of ground truth class $y_{\text{true}}$ have higher chance of reaching high-affinity. The evolutionary optimization thus promotes the label \textit{specificity} towards the ground truth. The solution denotes the data-label pairs of examples with high-affinity to the input, which we call plasma data. After the process, a majority vote of plasma data is used to make the class prediction. We refer readers to Section~\ref{sec: alg_detail} for more implementation details and Section~\ref{sec: perf_layer} for visualization. In short, static learning defenses seek to correct the predictions of current adversarial inputs and do not plan ahead for future attacks.

\subsection{Defense with adaptive learning}
Different from static learning, RAILS adaptive learning tries to use information from past attacks to harden the classifier to defend against future attacks. Hardening is done by leveraging another set of data - memory data generated during evolutionary optimization. Unlike plasma data, memory data is selected from examples with moderate-affinity to the input, which can rapidly adapt to new variants of the adversarial examples. This selection logic is based on maximizing coverage over future attack-space rather than optimizing for the current attack. Adaptive learning is a life-long learning process and can use hindsight to greatly enhance resilience of  $p_l^{y_{\text{true}}(\mathbf x)}(\mathbf x)$ to attacks. This paper will focus on static learning and single-stage adaptive learning that implements a single cycle of classifier hardening.


\subsection{A biological perspective} 

RAILS is inspired by and closely associated with the biological immune system. The architecture of the adaptive immune system ensures a robust response to foreign antigens to produce an effective antibody. Figure~\ref{fig: bio_comp} displays a comparison between the immune system workflow and the RAILS workflow. Both systems are composed of a four-step process. For example, RAILS emulates flocking from the immune system by initializing a population of candidates that provide \textit{diversity}, and emulates affinity maturation via an evolutionary optimization process to promote \textit{specificity}. Similar to the functions of plasma B cells and memory B cells generated in the immune system, RAILS generates plasma data for predictions of the present inputs (immune system defends antigens though generating plasma B cells) and generates memory data for the defense against future attacks (immune system continuously increases its degree of robustness through generating memory B cells). We refer readers to Appendix~\ref{sec: app1} for a table of correspondences between RAILS operands and mechanisms in the immune system. In addition, the learning patterns of RAILS and the immune system are closely aligned, as shown in Figure~\ref{fig: curves_bio_rails}.

\section{Details on RAILS Workflow}\label{sec: alg_detail}


Algorithm~\ref{RAILS} shows the four-step workflow of RAILS. We explain each step below.

\paragraph{Sensing.} This step performs an initial discrimination between adversarial and benign inputs to prevent the RAILS computation from becoming overwhelmed by false positives, i.e., only implementing the main steps of RAILS on suspicious inputs. While there are many outlier detection procedures that could be used for this step  \cite{feinman2017detecting,xu2017feature}, we can exploit the fact that the DNN and kNN applied on hidden layers will tend to make similar class predictions for benign inputs. Thus we propose using an cross-entropy measure to generate an {\it adversarial threat score} for each input $\mathbf x$. In the main text results, we skip the sensing stage since the major benefit from sensing is providing an initial detection. We refer readers to Appendix~\ref{sec: app5} for more details.

\paragraph{Flocking}
The initial population from each class needs to be selected with a certain degree of affinity measured using the hidden representations in order to satisfy our hypothesis, as illustrated in Figure~\ref{fig: curves_bio_rails}. By constructing class-wise k-Nearest Neighbor (kNN) architecture, we find the kNN that have the highest initial affinity score to the input data from each class and each selected layer. Mathematically, we select
\begin{align}\label{eq: knn}
    \begin{array}{ll}
\mathcal{N}_l^c = \{(\hat{\mathbf x}, y_c)|R_c(\hat{\mathbf x}) \le k, (\hat{\mathbf x}, y_c) \in \din_c\} \\ \text{Given}\\ A(f_l; \mathbf x_i^c, \mathbf x) \le A(f_l; \mathbf x_j^c, \mathbf x) \Leftarrow R_c(i) > R_c(j)\\ \forall c \in [C], l \in \mathcal{L},  \forall i, j \in [n_c],
    \end{array}
\end{align}
where $x$ is the input. $\mathcal{L}$ is the set of the selected layers. $\din_c$ is the training data from class $c$ and the size $|\din_c|=n_c$. $R_c: [n_c] \rightarrow [n_c]$ is a ranking function that sorts the indices based on the affinity score. In the adaptive learning context, if the memory database has been previously populated, flocking will select the nearest neighbors using both the training data and the memory data. The immune system leverages flocking step to find initial B cells and form temporary structures for affinity maturation \cite{de2015dynamics}. Note that in RAILS, the kNN sets $\mathcal{N}_l^c $ are \textit{constructed independently} for each class, thereby ensuring that every class is \textit{fairly represented} in the initial population.

\paragraph{Affinity maturation (evolutionary optimization)} As flocking brings diversity to the label distribution of the initial population, the affinity maturation step, in contrast, promotes specificity towards the ground truth class. Here we use evolutionary optimization to generate new examples (offspring) from the existing examples (parents) in the population. The evolution happens within each class \textit{independently}, and new generated examples from different classes are not affected by one another before the consensus stage. The first generation parents in each class are the $K$ nearest neighbors found by \eqref{eq: knn} in the flocking step, where $K$ is the number of nearest neighbors. Given a total population size $TC$, the $0$-th generation is obtained by copying each nearest neighbor $T/K$ times with random mutations. Given the population of class $c$ in the $(g-1)$-st generation $\mathcal{P}_c^{(g-1)}=[\mathbf x_c(1), \mathbf x_c(2), \cdots, \mathbf x_c(T)] \in \mathbb{R}^{d \times T}$, the candidates for the $g$-th generation are selected as
\begin{align}\label{eq: random}
    \begin{array}{ll}
\mathcal{\hat{P}}_c^{(g-1)} = \mathcal{P}_c^{(g-1)} Z_c^{(g-1)},
    \end{array}
\end{align}
where $\mathcal{\hat{P}}_c^{(g-1)}$ denotes the set of randomly selected reproductions of the population at the previous generation $g-1$. These are  computed before applying 
mutation and crossover operations to populate the $g$-th generation $\mathcal{P}_c^{(g)}$. $Z_c^{(g-1)} \in \mathbb{R}^{T \times T}$ is a binary selection matrix whose columns are independent and identically distributed draws from  Mult$(1, \mathbf P_c)$, the multinomial distribution with probability vector $\mathbf P_c\in [0,1]^T$. The process can also be viewed as creating new nodes from existing nodes in a Preferential Attachment (PA) evolutionary graph \cite{barabasi1999emergence}, where the details can be viewed in Appendix~\ref{sec: app4}. After \textit{selection}, RAILS generates new examples through the operations of \textit{mutation} and \textit{cross-over}, which will be discussed in more detail later. After new examples are generated, RAILS calculates each example’s affinity relative to the input. The new examples are associated with labels that are inherited from their parents, which always come from the same class. According to our hypothesis in Section \ref{sec: RAILS}, examples inherited from parents of the ground truth class $y_{\text{true}}$ have a higher chance of reaching high-affinity, and thereby the population members with high-affinity are concentrating about the input's true class.

\paragraph{Consensus} Consensus is responsible for the final selection and predictions. In this step, RAILS selects generated examples with high-affinity scores to be plasma data, and examples with moderate-affinity scores are saved as memory data. The selection is based on a ranking function. 
\begin{align}\label{eq: aff_mat}
    \begin{array}{ll}
S_{\text{opt}} = \{(\Tilde{\mathbf x}, \Tilde{y})|R_g(\Tilde{\mathbf x}) \le \gamma|\mathcal{P}^{(G)}|, (\Tilde{\mathbf x}, \Tilde{y}) \in \mathcal{P}^{(G)}\},
    \end{array}
\end{align}
where $R_g: [|\mathcal{P}^{(G)}|] \rightarrow [|\mathcal{P}^{(G)}|]$ is the same ranking function as $R_c$ except that the domain is a set having cardinality equal to that of the final population $\mathcal{P}^{(G)}$. $\gamma$ is a proportionality parameter and is selected as $0.05$ and $0.25$ for plasma data and memory data, respectively. Note that the memory data can be selected in each generation. 
For simplicity, we select memory data only in last generation. Memory data will be saved in the secondary database and used for model hardening. 

Given that all examples in the population are associated with a label inherited from their parents, RAILS uses majority voting of the plasma data for prediction of the class label of $\mathbf x$.

\begin{algorithm}[h]
\caption{Robust Adversarial Immune-inspired Learning System (RAILS)}
\label{RAILS}
\begin{algorithmic}[1]
\REQUIRE Test data point $\mathbf x$; Training dataset $\din_{tr} = \{\din_{1}, \din_{2},\cdots,\din_{C}\}$; Number of Classes $C$; Model $M$ with feature mapping $f_l(\cdot)$, $l \in \mathcal{L}$; Affinity function $A$.\\
\noindent{\textbf{First Step: Sensing}}
\STATE{Check the threat score given by an outlier detection strategy to detect the threat of $x$.}\\
\noindent{\textbf{Second Step: Flocking}}
\FOR{$c = 1, 2, \dots, C$}
\STATE{In each layer $l \in \mathcal{L}$, find the k-nearest neighbors $\mathcal{N}_l^c$ of $\mathbf x$ in $\din_{c}$ by ranking the affinity score.} 
\ENDFOR\\
\noindent{\textbf{Third Step: Affinity Maturation}}

\STATE{\textbf{For} each layer $l \in \mathcal{L}$, \textbf{do}}
\STATE{Generate $\mathcal{P}_c^{(0)}$ through mutating each of $\mathbf x^{\prime} \in \mathcal{N}_l^c$ $T/K$ times, $\forall c \in [C]$.}
\FOR{$g = 1, 2, \dots, G$ }
\FOR{$t = 1, 2, \dots, T$}
\STATE{Select data-label pairs $(\mathbf x_c, y_c)$, $(\mathbf x_c^{\prime}, y_c)$ from $\mathcal{P}_c^{(g-1)}$ based on $\mathbf P_c^{(g-1)}$.}
\STATE{$\mathbf x_{\text{os}} = Mutation \big (Crossover(\mathbf x_c,\mathbf x_{c}^{\prime})\big)$; $\mathcal{P}_c^{(g)} \longleftarrow (\mathbf x_{\text{os}}, y_c)$.}
\ENDFOR
\ENDFOR
\STATE{Calculate the affinity score $A(f_{l}; \mathcal{P}^{(G)}, \mathbf x), \forall c \in [C]$ given $\mathcal{P}^{(G)}=\mathcal{P}_1^{(G)}\bigcup \cdots \bigcup \mathcal{P}_C^{(G)}$.}
\STATE{\textbf{end For}}\\
\noindent{\textbf{Fourth Step: Consensus}}
\STATE{Select the top $5\%$ as plasma data $S_{p}^l$ and the top $25\%$ as memory data $S_{m}^l$ based on the affinity scores, $\forall l \in \mathcal{L}$; Obtain the prediction $y$ of $\mathbf x$ using the majority vote of the plasma data.}
\STATE {\bfseries Output:} $y$, the memory data $S_{m} = \{S_{m}^1, S_{m}^2,\cdots,S_{m}^{|\mathcal{L}|}\}$
\end{algorithmic}
\end{algorithm}

\paragraph{Computational cost} 
The computational cost of RAILS is dominated by the flocking and affinity maturation stage. kNN structure construction in flocking is a fixed setup cost that can be handled off-line with fast approximate kNN approximation  \cite{bewley2013advantages,rajaraman2011mining}. There are three strategies for reducing  the computational cost of the affinity maturation stage. 
First, the evolutionary optimization can be replaced by a mean field approximation. Second, parallelization can be used to accelerate the computations since each sample is generated and utilized separately. Third, one can use a more stringent false positive threshold in the sensing step, thereby reducing the number of false positives resulting in a reduction in the downstream computational burden. More discussion can be viewed in Appendix~\ref{sec: app4}. 

\paragraph{Operations in the evolutionary optimization.}
Three operations support the creation of new examples: selection, cross-over, and mutation. 
The \textit{selection} operation is shown in \eqref{eq: random}. We compute the selection probability for each candidate through a softmax function.
\begin{align}\label{eq: prob}
    \begin{array}{ll}
\mathbf P(\mathbf x_i) &= Softmax(A(f_{l}; \mathbf x_i, \mathbf x)/\tau) \\& = \frac{\exp{(A(f_{l}; \mathbf x_i, \mathbf x)/\tau)}}{\sum_{\mathbf x_j \in S}\exp{(A(f_{l}; \mathbf x_j, \mathbf x)/\tau)}},
    \end{array}
\end{align}
where $S$ is the set of data points and $x_i \in S$. $\tau>0$ is the sampling temperature that controls sharpness of the softmax operation. Given the selection probability $\mathbf P$, defined on the current generation in \eqref{eq: prob},  the candidate set $\{(\mathbf x_i,y_i)\}_{i=1}^T$ for the next generation is randomly drawn (with replacement). 

The \textit{cross-over} operator combines two parents $\mathbf x_{c}$ and $\mathbf x_{c}^{\prime}$ from the same class, and generates new offspring by randomly selecting each of its elements (pixels) from the corresponding element of either parent. Mathematically,
\begin{align}\label{eq: crossover}
    \begin{array}{ll}
&\mathbf x_{\text{os}}^{\prime} = Crossover(\mathbf x_{c}, \mathbf x_{c}^{\prime}) =\\& \left \{ \begin{array}{rcl}
\mathbf x_{c}^{(i)} & \mbox{with prob}~ \frac{A(f_{l}; \mathbf x_{c}, \mathbf x)}{A(f_{l}; \mathbf x_{c}, \mathbf x)+A(f_{l}; \mathbf x_{c}^{\prime}, \mathbf x)}, \\ \mathbf x_{c}^{\prime{(i)}} & \text{with prob}~ \frac{A(f_{l}; \mathbf x_{c}^{\prime}, \mathbf x)}{A(f_{l}; \mathbf x_{c}, \mathbf x)+A(f_{l}; \mathbf x_{c}^{\prime}, \mathbf x)}
\end{array} \right. \forall i \in [d],
    \end{array}
\end{align}
where $i$ represents the $i$-th entry of the example and $d$ is the dimension of the example. The \textit{mutation} operation randomly and independently  mutates an offspring with probability $\rho$, adding uniformly distributed noise in the range $[-\delta_{\text{max}},-\delta_{\text{min}}] \cup [\delta_{\text{min}},\delta_{\text{max}}]$. The resulting perturbation vector is subsequently clipped to satisfy the domain constraint that examples lie in $[0,1]^d$.
\begin{align}\label{eq: mutation}
    \begin{array}{ll}
& \mathbf x_{\text{os}} = Mutation(\mathbf x_{\text{os}}^{\prime}) = \text{Clip}_{[0,1]}\big(\mathbf x_{\text{os}}^{\prime} +\\& \boldsymbol{1}_{[Bernoulli(\rho)]}  \mathbf u([-\delta_{\text{max}},-\delta_{\text{min}}] \cup [\delta_{\text{min}},\delta_{\text{max}}])\big),
    \end{array}
\end{align}
where $\boldsymbol{1}_{[Bernoulli(\rho)]}$ takes value $1$ with probability $\rho$ and value $0$ with probability $1-\rho$. $\mathbf u([-\delta_{\text{max}},-\delta_{\text{min}}] \cup [\delta_{\text{min}},\delta_{\text{max}}])$ is the vector in $\mathbb R^d$ having i.i.d. entries drawn from the punctured uniform distribution $\mathcal{U}([-\delta_{\text{max}},-\delta_{\text{min}}] \cup [\delta_{\text{min}},\delta_{\text{max}}])$. $\text{Clip}_{[0,1]}(\mathbf x)$ is equivalent to $\max(0, \min(\mathbf x, 1))$.

\section{Experimental results}\label{sec: experiments}
We conduct experiments in the context of image classification using several benchmark image classification datasets. We compare RAILS with Convolutional Neural Network (CNN) Classification and Deep k-Nearest Neighbors (DkNN) Classification \cite{papernot2018deep} on the MNIST \cite{lecun1998gradient}, SVHN \cite{netzer2011reading}, CIFAR-10 and CIFAR-100 \cite{Krizhevsky2009learning} datasets. We test our framework using a four-convolutional-layer neural network for MNIST, VGG16 \cite{SZ2014} for the SVHN dataset, and adversarially trained VGG16 for the CIFAR-10 and CIFAR-100 datasets. We refer readers to Appendix~\ref{sec: app2} for more details of the datasets, models, and parameter selection. In addition to the benign test examples, we also generate an equal number of adversarial examples generated from 
adversarial attacks using MNIST, SVHN, CIFAR-10, and CIFAR-100 data. The attack strength is $\epsilon = 40/60$ for MNIST, and $\epsilon = 8$ for SVHN, CIFAR-10, CIFAR-100 by default. Performance comparisons are based on standard accuracy (SA) evaluated using benign (unperturbed) test examples and robust accuracy (RA) evaluated using the adversarial (perturbed) test examples. Besides the default $\ell_\infty$ norm PGD attack \cite{madry17}, we also implement seven other attacks (Figure~\ref{fig: overview_result} and Table~\ref{tab: acc_various_attack}).

\subsection{Performance in single layers}\label{sec: perf_layer}

\begin{figure}[h]
  \centering
  \begin{adjustbox}{max width=0.43\textwidth }
  \begin{tabular}{@{\hskip 0.00in}c  @{\hskip 0.00in}c @{\hskip 0.02in} @{\hskip 0.02in} c @{\hskip 0.02in} @{\hskip 0.02in}c }
 \begin{tabular}{@{}c@{}}  
\vspace*{0.005in}\\
\rotatebox{90}{\parbox{17em}{\centering \Large \textbf{Adv examples ($\epsilon=60$)}}}
 \\
\end{tabular} 
&
\begin{tabular}{@{\hskip 0.02in}c@{\hskip 0.02in}}
     \begin{tabular}{@{\hskip 0.00in}c@{\hskip 0.00in}}
     \parbox{10em}{\centering \Large Conv1}  
    \end{tabular} 
    \\
 \parbox[c]{20em}{\includegraphics[width=20em]{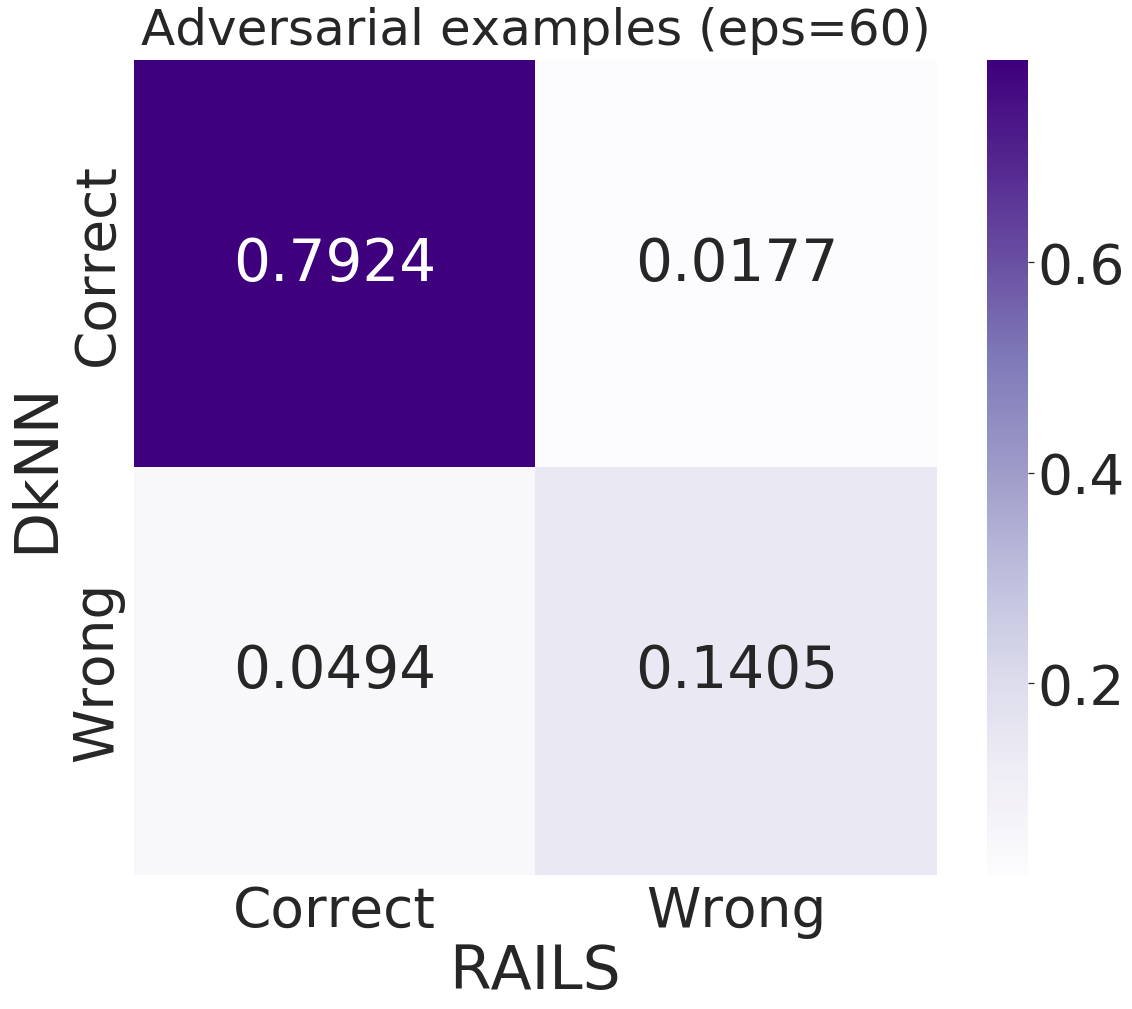}} 
\end{tabular}
&
 \begin{tabular}{@{\hskip 0.02in}c@{\hskip 0.02in}}
      \begin{tabular}{@{\hskip 0.00in}c@{\hskip 0.00in}}
     \parbox{10em}{\centering \Large Conv2}
    \end{tabular} 
    \\
 \parbox[c]{20em}{\includegraphics[width=20em]{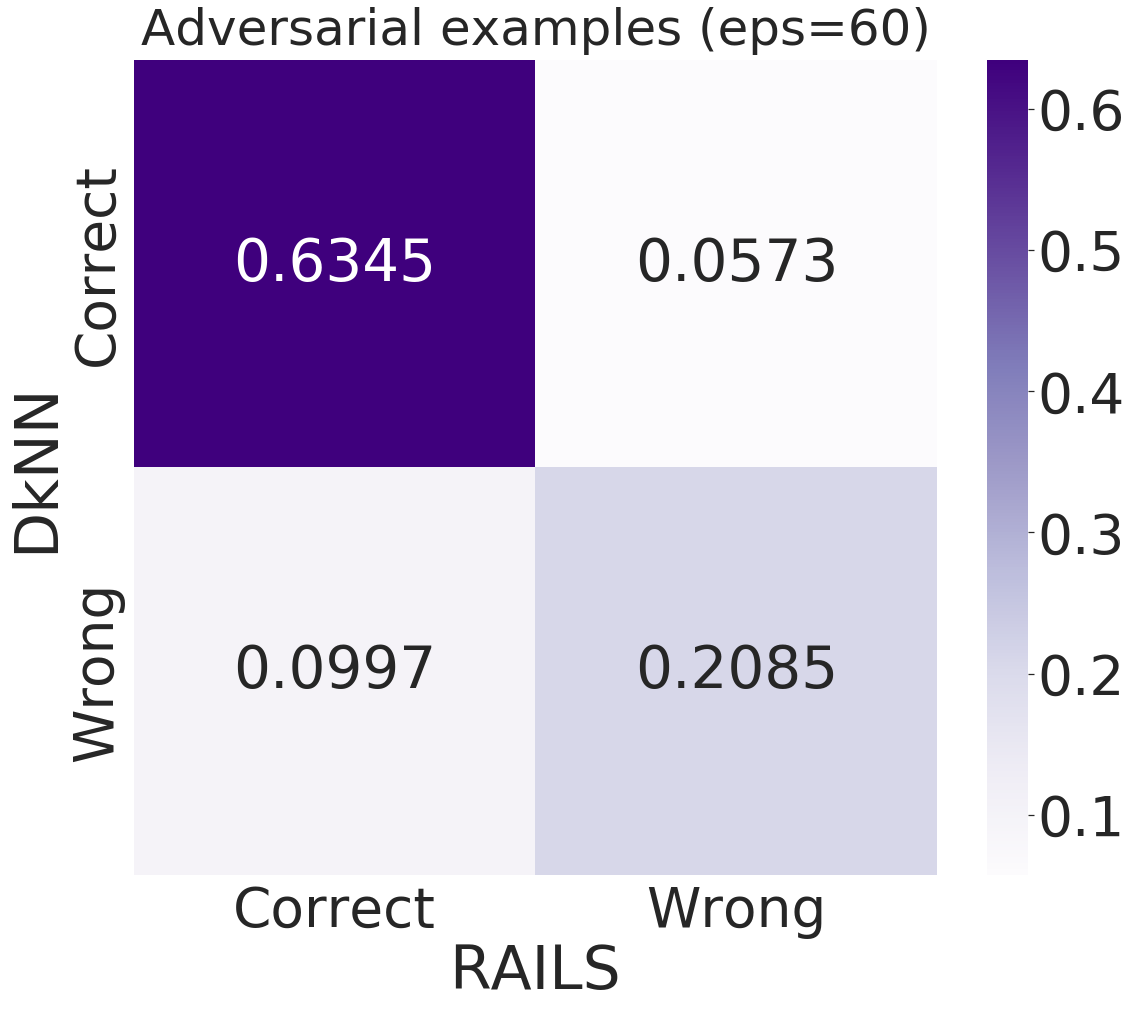}} 
\end{tabular}

\end{tabular}
  \end{adjustbox}
    \caption{\textbf{RAILS has fewer incorrect predictions for those data that DkNN gets wrong.} Confusion Matrices of adversarial examples classification in Conv1 and Conv2 (RAILS vs. DkNN).}
  \label{fig: conf_mat}
\end{figure}
We first test RAILS in a single layer of the CNN model and compare the obtained accuracy with the results from the DkNN. Table~\ref{tab1: acc_layer} shows the comparisons in the input layer, the first convolutional layer (Conv1), and the second convolutional layer (Conv2) on MNIST. One can see that for both standard accuracy and robust accuracy, RAILS performs better than the DkNN in the hidden layers and achieve better results in the input layer. The input layer results indicate that RAILS can also outperform supervised learning methods like kNN. The confusion matrices in Figure~\ref{fig: conf_mat} show that RAILS has fewer incorrect predictions for those data that DkNN gets wrong. Each value in Figure~\ref{fig: conf_mat} represents the percentage of intersections of RAILS (correct or wrong) and DkNN (correct or wrong).

\begin{table}[h]
\begin{center}
\caption{\textbf{RAILS outperforms DkNN on single layers.} Standard Accuracy (SA)/Robust Accuracy (RA) performance of RAILS versus DkNN in single layer (MNIST).}
\label{tab1: acc_layer}
\resizebox{0.46\textwidth}{!}{
\begin{tabular}{lc||c|c|c}
\hline
\hline
 &  &  Input&  Conv1& Conv2 \\
\hline 
SA &  \bf{RAILS}& \bf{97.53\%} & \bf{97.77\%} & \bf{97.78\%} \\
 & DkNN & 96.88\%  & 97.4\% & 97.42\% \\
\hline
RA & \bf{RAILS}& \bf{93.78\%}  & \bf{92.56\%} & \bf{89.29\%} \\
($\epsilon=40$) & DkNN& 91.81\%  & 90.84\% & 88.26\% \\
\hline 
RA &  \bf{RAILS}& \bf{88.83\%} & \bf{84.18\%} & \bf{73.42\%} \\
($\epsilon=60$) &  DkNN& 85.54\% & 81.01\% & 69.18\% \\

\hline
\hline
\end{tabular}}
\end{center}
\end{table}

\begin{figure*}[h]
  \centering
  \begin{adjustbox}{max width=0.9\textwidth }
    \begin{tabular}{@{\hskip 0.00in}c  @{\hskip 0.02in} c @{\hskip 0.02in} @{\hskip 0.02in} c }

 \begin{tabular}{@{}c@{}}  
\rotatebox{90}{\parbox{10em}{\centering \scriptsize\textbf{{\begin{tabular}[c]{@{}c@{}}  Example 1 \\ RAILS: bird \\ DkNN: cat  \end{tabular}}{}}}} \\

\rotatebox{90}{\parbox{6.5em}{\raggedleft \scriptsize \textbf{{\begin{tabular}[c]{@{}c@{}} Example 2 \\ RAILS: bird \\ DkNN: horse \end{tabular}}{}}}}
\\
\end{tabular} 
&
 \begin{tabular}{@{\hskip 0.02in}c@{\hskip 0.02in}c@{\hskip 0.02in}}
 \begin{tabular}{@{\hskip 0.02in}c@{\hskip 0.02in}c@{\hskip 0.02in}c@{\hskip 0.02in}c@{\hskip 0.02in}}
\begin{tabular}{@{\hskip 0.02in}c@{\hskip 0.02in}}
\\
 \parbox[c]{10em}{              }{\includegraphics[width=6em]{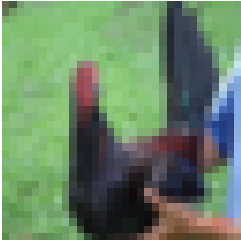}}
 \\
  \parbox[c]{10em}{              }{\includegraphics[width=6em]{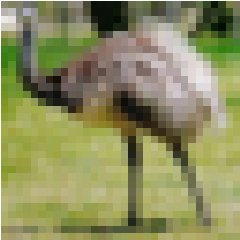}} 
\end{tabular}

 \begin{tabular}{@{\hskip 0.02in}c@{\hskip 0.02in}}
\\
\parbox[c]{10em}{\includegraphics[trim=5 0 0 0,clip,width=9.5em]{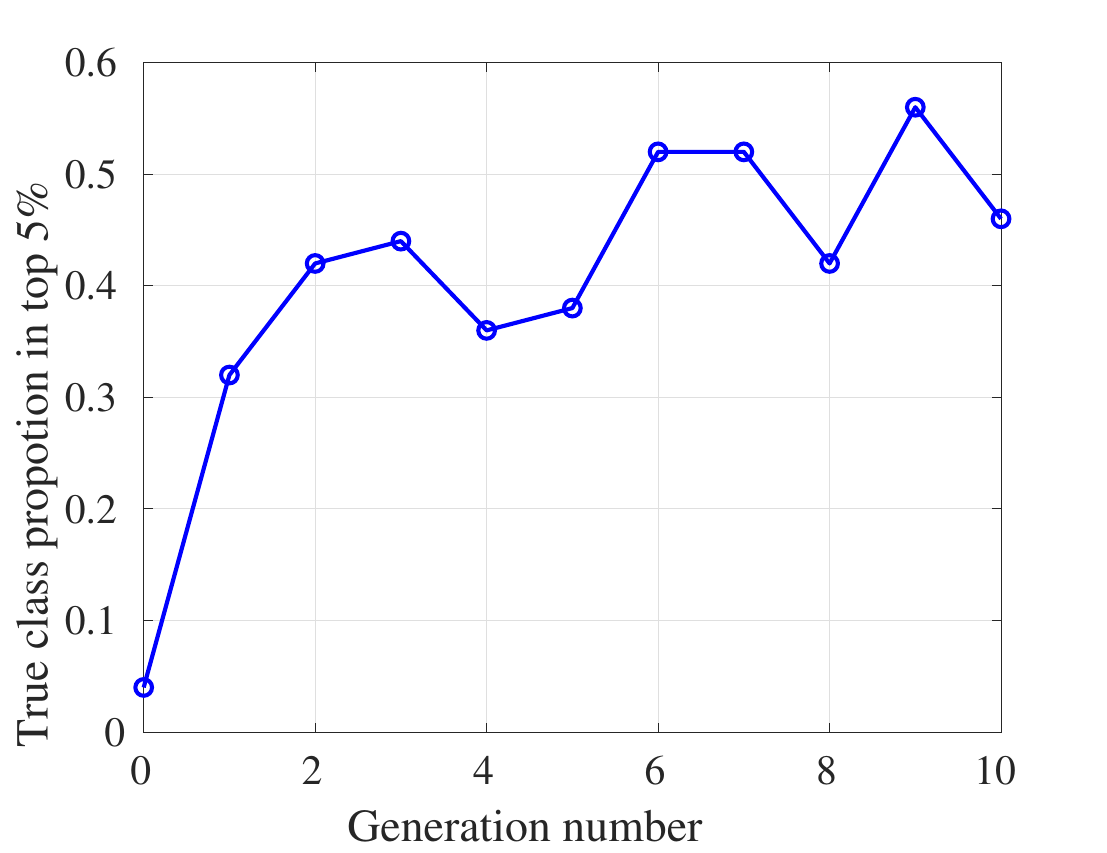}}
 \\
  \parbox[c]{10em}{\includegraphics[trim=5 0 0 0,clip,width=9.5em]{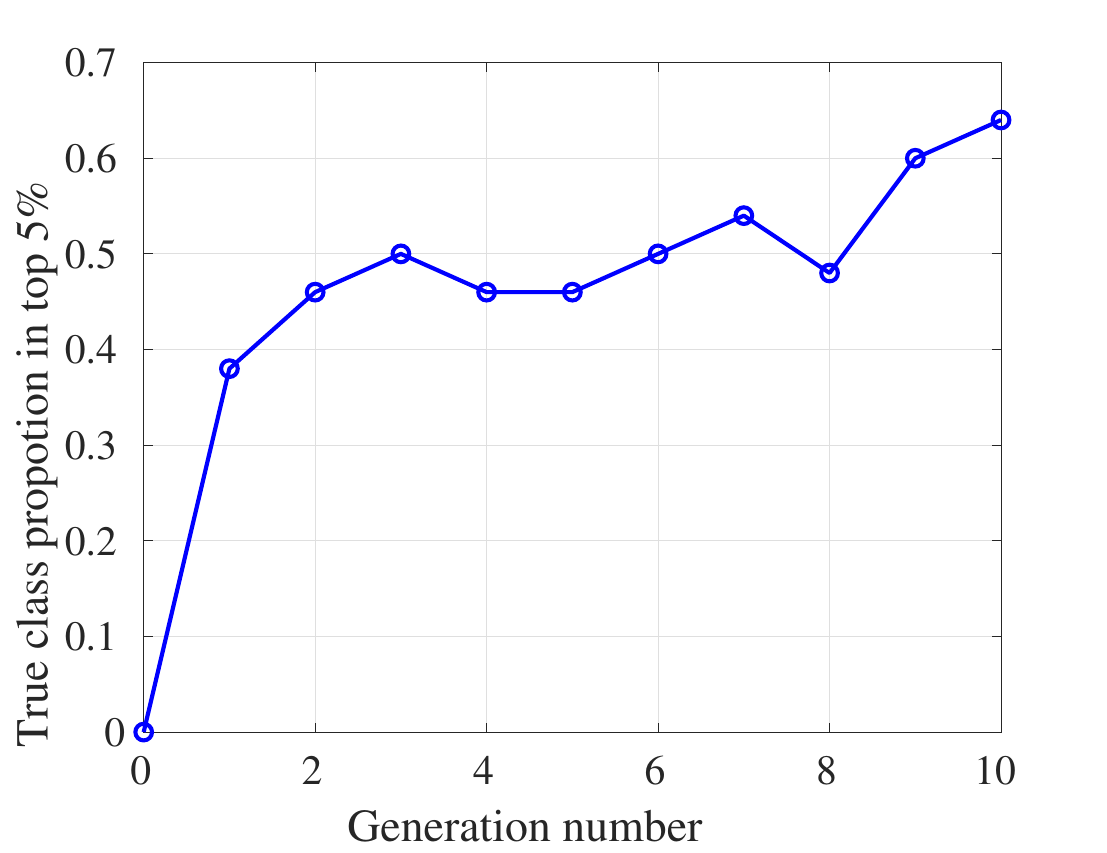}} 

\end{tabular}

 \begin{tabular}{@{\hskip 0.02in}c@{\hskip 0.02in}}
\\
 \parbox[c]{10em}{\includegraphics[trim=5 0 0 0,clip,width=9.5em]{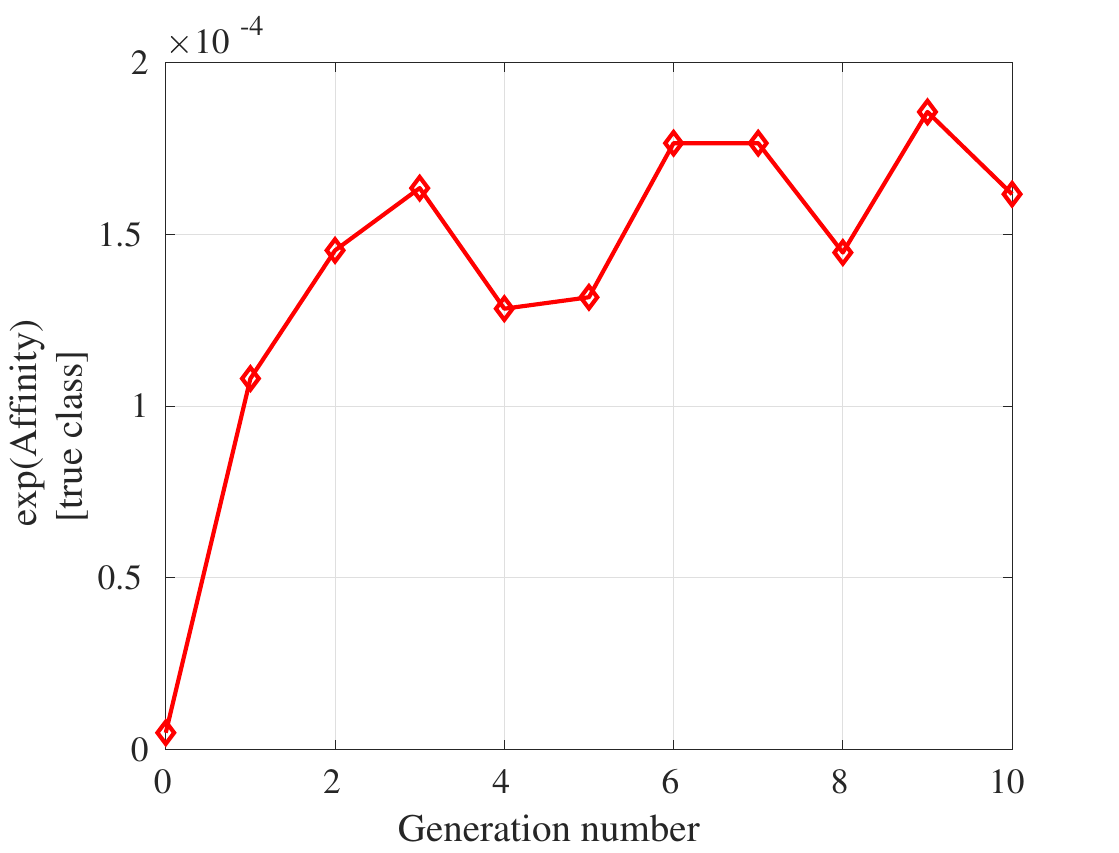}}
  \\
 \parbox[c]{10em}{\includegraphics[trim=5 0 0 0,clip,width=9.5em]{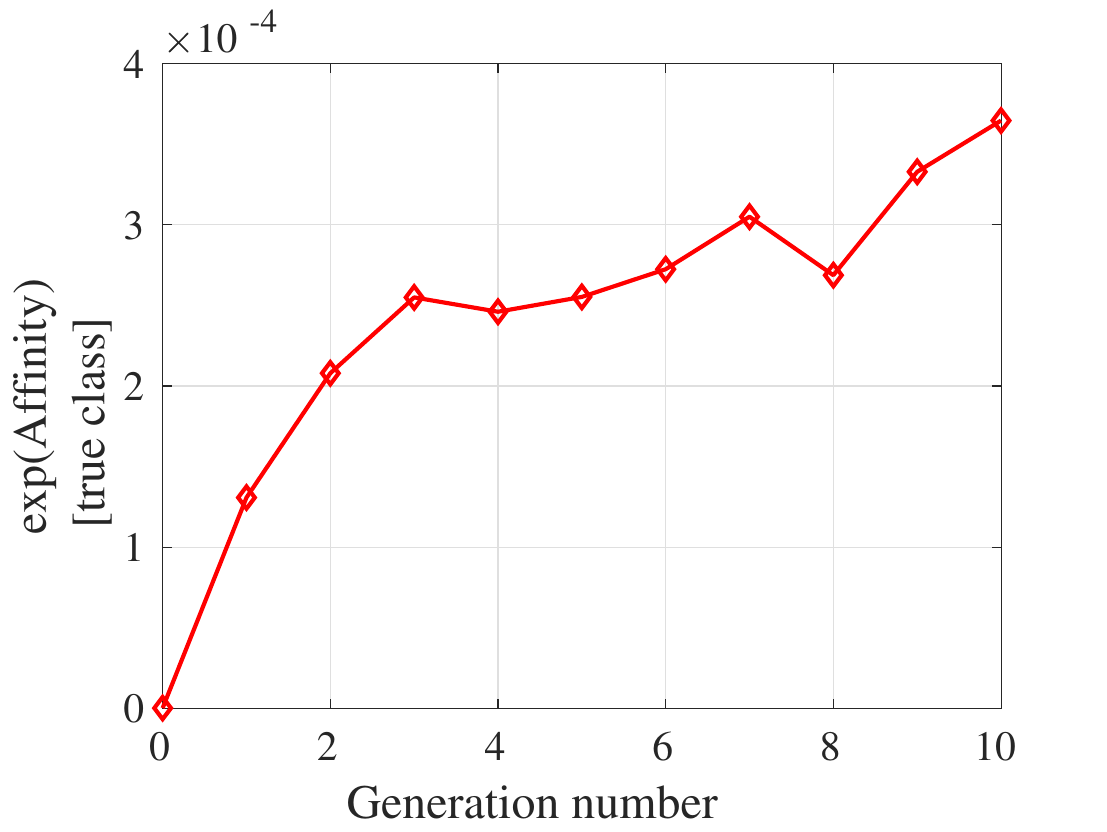}}

\end{tabular}

\end{tabular}
\end{tabular} 

\end{tabular}
  \end{adjustbox}
    \caption{Proportion and affinity of the population from the ground truth class of input with respect to the generation number (RAILS on CIFAR-10). We plot all curves by selecting data points with affinity in the top $5\%$ of all classes' data points in each generation. (1) Second column: Data from the true class occupies the majority of the population when the generation number increases (2) Third column: Affinity maturation over multiple generations produces increased affinity (after a temporary decrease in the searching phase) within the true class.}
  \label{fig: curves_running}
\end{figure*}

\subsection{RAILS learning process}\label{sec: vis}
Flocking provides a balanced initial population while affinity maturation within RAILS creates new examples in each generation. To better understand the capability of RAILS, we can visualize the changes of some key indices during runtime.

\begin{figure}[h]
\centerline{\includegraphics[width=.46\textwidth]{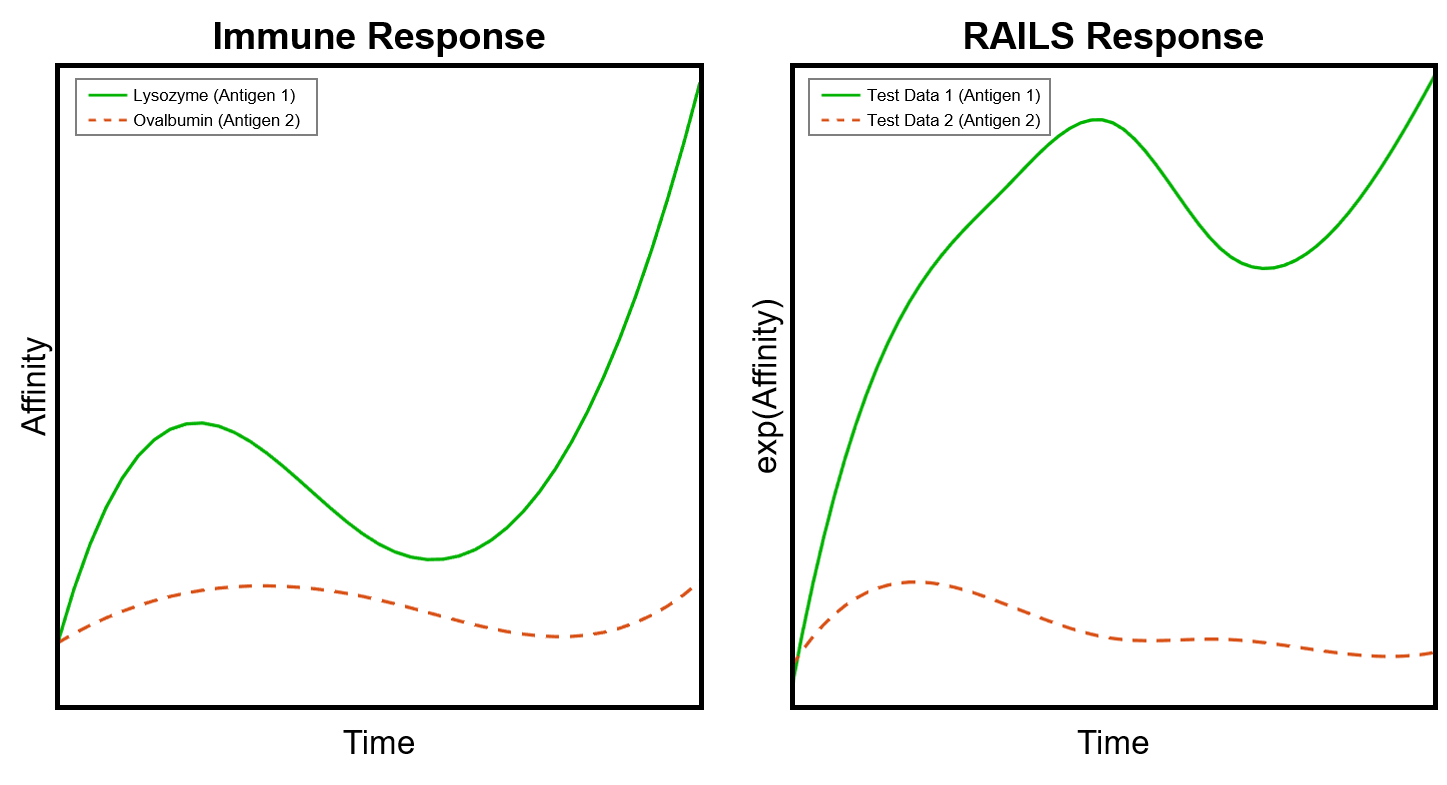}}
\caption{\textbf{RAILS emulates the learning patterns of the biological adaptive immune response.} Correspondence between learning curves of a natural immune system \textit{in vitro} experiment (left) and the RAILS computational experiment (right). Adversarial input in RAILS corresponds to antigen in the immune system. When the initial candidates are selected based on the input (the green lines), RAILS and the immune system can both jump out of local optimal and find the correct solution. When the candidates are selected based on a different input (the red dashed lines), neither responses converge.}
\label{fig: curves_bio_rails}
\end{figure}

\begin{figure*}[h]
\centerline{\includegraphics[trim=0 0 0 0,clip,width=0.95\linewidth]{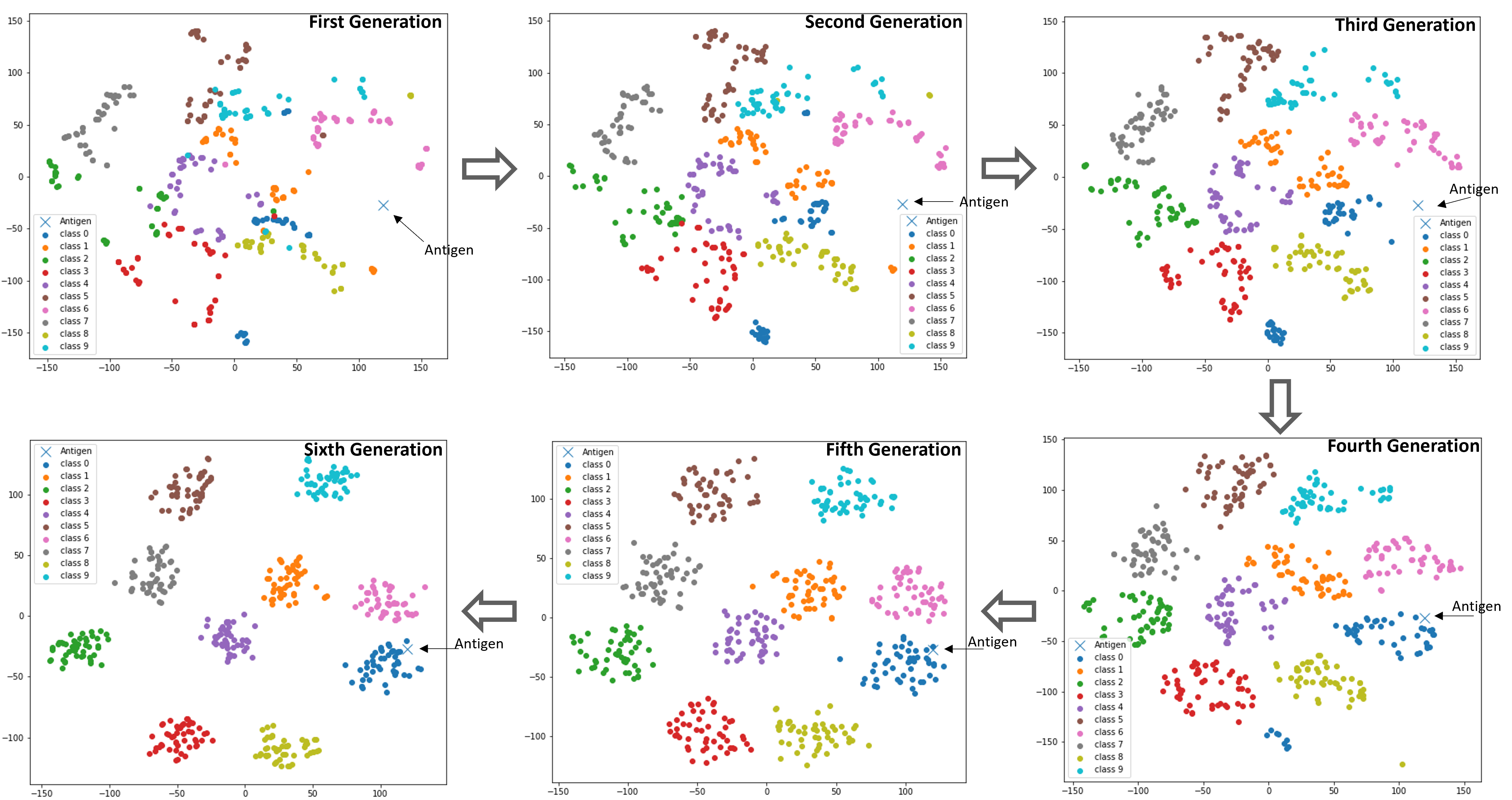}}
\caption{\textbf{Visualization of the evolution of B-cell feature samples over several generations of the the RAILS affinity maturation process.} Feature representations of an adversarial CIFAR-10 input (antigen) and the RAILS selected feature populations of B-cells are mapped to a two-dimensional space (t-SNE). The random initialization of the population displays the B-cells as uniformly distributed over feature space.  After six generations the affinity maturation process produces B-cells that  cluster around the antigen and correctly identify its true class.
}
\label{fig: rails_tsne}
\end{figure*}

\paragraph{Visualization of RAILS evolutionary process.}
Picking the top $5\%$ data points with the highest affinity in each generation, Figure~\ref{fig: curves_running} shows the evolution over ten generations of RAILS samples of the population (B-cells) proportion and (exponentiated) affinity relative to two clean (non-adversarial) input examples taken  
from CIFAR-10. RAILS makes the correct ``bird'' predictions while the DkNN makes incorrect predictions for both examples. The second column depicts the proportion of the true class in the selected population of each generation. Data from the true class occupies the majority of the population when the generation number increases, which indicates that RAILS can obtain a correct prediction and a high confidence score simultaneously. Meanwhile, affinity maturation over multiple generations yields increasing affinity within the true class, as shown in the third column. To visualize changes in feature distribution during the affinity maturation stage, we show in Figure~\ref{fig: rails_tsne} the  two-dimensional t-distributed stochastic neighbor embedding (t-SNE) of the feature representations of adversarial CIFAR-10 inputs (antigens) and the associated populations (B-cells). The features shown in the figure  are those of convolutional layer three, and are representative of the feature behavior at other layers.  
As shown in Figure~\ref{fig: rails_tsne}, the antigen is misclassified and B-cells are uniformly spread over the feature space at the beginning of the affinity maturation. As the affinity maturation process progresses, the antigen's ground truth class B-cell population (colored in blue) converges to a cluster that covers the antigen.

\paragraph{In-vitro B-cell experiment confirms RAILS emulation}

To demonstrate that the proposed RAILS computational system captures important properties of the actual (in-vitro) immune system we compare the learning curve of RAILS to the learning curve of B-cell antigen recognition (see Appendix~\ref{sec: app3} for a description of the biological experiment we performed). For the biological experiment the measured affinity between a population of actual B-cells and an antigen is obtained experimentally over time (several hours).   For RAILS  each test input (potentially the adversarial example) is treated as an antigen and the affinity is computed as RAILS iterates over time. Figure~\ref{fig: curves_bio_rails} shows that both the in-vitro immune system and RAILS have similar learning patterns. One can also see that the affinity increases again after the decrease, indicating both the immune system and RAILS can escape from a local optimal under strong attacks. The difference between the green and red curves is that the initial population for the red curve is found based on another test input (antigen), which has lower correlation to the current input (antigen). The non-convergence of the red curve indicates that the initial population should be selected close to the input, and the flocking using kNN search emulated the natural flocking process. We refer readers to Appendix~\ref{sec: app3} for more details.

\subsection{Overall performances in different scenarios}

\paragraph{Defense against PGD attack for CIFAR-10 and SVHN} We compare RAILS with CNN and DkNN in terms of standard accuracy (SA) and robust accuracy (RA). The results are shown in Table~\ref{tab2: acc_overall}. On MNIST with $\epsilon=60$, one can see that RAILS delivers a $5.62\%$ improvement in RA over DkNN without appreciable loss of SA. On CIFAR-10 (SVHN), RAILS leads to $10.32\%$ ($12.5\%$) and $19.4\%$ ($46\%$) robust accuracy improvements compared to DkNN and CNN, respectively. We refer readers to Appendix~\ref{sec: app6} for more results under different strengths and types of attacks. Note that there is no competitive relationship between RAILS and robust training since RAILS is a general method that can improve any models' robustness, even a robust trained model.

\begin{table}[h]
\begin{center}
\caption{\textbf{RAILS achieves higher robust accuracy (RA) at small cost of standard accuracy (SA)} for all three datasets (MNIST, SVHN and CIFAR-10) as compared to CNN and DkNN.
}
\label{tab2: acc_overall}
\resizebox{0.43\textwidth}{!}{
\begin{tabular}{lc|c|c}
\hline
\hline 
& & SA & RA \\
\hline 
MNIST &\bf{RAILS (ours)} & 97.95\% & \bf{76.67\%}  \\

($\epsilon=60$) &CNN & \bf{99.16\%} & 1.01\% \\

& DkNN & 97.99\% & 71.05\%   \\
\hline 
SVHN  &\bf{RAILS (ours)}  & 90.62\% & \bf{48.26\%} \\

($\epsilon=8$) &CNN & \bf{94.55\%} & 1.66\% \\

&DkNN & 93.18\% & 35.7\% \\
\hline 
CIFAR-10 &\bf{RAILS (ours)} & 82\% & \bf{52.01\%}\\

($\epsilon=8$) &CNN & \bf{87.26\%} & 32.57\% \\

&DkNN & 86.63\% & 41.69\% \\

\hline
\hline
\end{tabular}}
\end{center}
\end{table}

\paragraph{Defense against eight different attack types}
Here we show the results of RAILS defending against eight types of attacks: (1,2) $\ell_\infty$-PGD attack and $\ell_2$-PGD attack \cite{madry17} (3) Fast Gradient Sign Method (FGSM) \cite{goodfellow2014explaining}, a fast alternative version of PGD (4) Square Attack (Sq-Attack) \cite{andriushchenko2020square}, a type of score-based black-box attack (5) Boundary Attack \cite{brendel2017decision}, a type of decision-based black-box attack (6) AutoAttack \cite{croce2020reliable}, a multi-level white-box attack (7) Adversarial Patch (Adv-P) \cite{brown2017adversarial}, an attack with unified perturbations across different inputs (8) a (customized) ASK-Attack that is directly applied on the flocking step \cite{wang2021ask}. We refer readers to Appendix~\ref{sec: app2} for details of the threat models. The results of RAILS defending against these attacks can be viewed in Figure~\ref{fig: overview_result} and Table~\ref{tab: acc_various_attack}. On CIFAR-10, RAILS improves the robust accuracy of CNN (DkNN) on $\ell_\infty$-PGD/$\ell_2$-PGD/FGSM/Sq-Attack/Boundary Attack/AutoAttack/Adv-P/ASK-Attack by $19.43\%/14.8\%/11.18\%/11.5\%/32.79\%/22.58\%/22.36\%-/11.81\%$ ($10.31\%/10.46\%/6.24\%/3.2\%/6.4\%/11.07\%-/10.8\%/7.7\%$). More experimental results can be found in Appendix~\ref{sec: app6}.

\begin{table}[h]
\begin{center}
\caption{\textbf{RAILS achieves higher robust accuracy (RA)} under eight types of attacks as compared to CNN and DkNN.
}
\label{tab: acc_various_attack}
\resizebox{0.48\textwidth}{!}{
\begin{tabular}{l|c|c|c}
\hline
\hline
& \textbf{RAILS} & DkNN & CNN \\
\hline 
$\ell_\infty$-PGD ($\epsilon=8$) & \bf{52.01\%} & 41.69\% & 32.57\%  \\
\hline
$\ell_2$-PGD ($\epsilon=127.5$) & \bf{35.1\%} & 24.64\% & 20.3\% \\
\hline
FGSM ($\epsilon=8$) & \textbf{59.7\%} & 53.46\% & 48.52\% \\
\hline
Sq-Attack ($\epsilon=20$) & \textbf{74.5\%} & 71.3\% & 53.7\% \\
\hline
Boundary Attack ($\ell_2$)  & $\bf{70.6\%}$  & $64.2\%$  & $37.81\%$ \\
\hline 
AutoAttack ($\epsilon=8$) & $\bf{52.84\%}$  & $41.77\%$  & $30.26\%$ \\
\hline
Adv-P (ratio$=0.1$) & \textbf{53.5\%} & 42.7\% & 31.14\% \\
\hline
ASK-Attack ($\epsilon=8$) & \textbf{45.5\%} & 37.8\% & 34.21\% \\
\hline
\hline
\end{tabular}}
\end{center}
\end{table}

We further evaluate RAILS on CIFAR-100 under the three most powerful attacks taken from Table~\ref{tab: acc_various_attack}: $\ell_\infty$-PGD; AutoAttack; and Boundary Attack. The standard accuracy of RAILS, DkNN, and CNN are $61.03\%$, $62.92\%$, and $65.57\%$. The results of defending against the three types of attacks on CIFAR-100 are shown in Table~\ref{tab: acc_cifar100}. One can see that RAILS outperforms adversarially trained CNN and DkNN.

\begin{table}[h]
\begin{center}
\caption{\textbf{RAILS achieves higher robust accuracy (RA)} than DkNN and CNN on CIFAR-100 under the three attacks.
}
\label{tab: acc_cifar100}
\resizebox{0.45\textwidth}{!}{
\begin{tabular}{l|c|c|c}
\hline
\hline
& \textbf{RAILS} & DkNN & CNN \\
\hline 
$\ell_\infty$-PGD ($\epsilon=8$) & \bf{41.35\%} & 32.96\% & $23.7\%$  \\
\hline 
AutoAttack ($\epsilon=8$) & $\bf{42.84\%}$  & $32.86\%$  & $25.63\%$ \\
\hline
Boundary Attack ($\ell_2$)  & $\bf{53.6\%}$  & $49.51\%$  & $29.1\%$ \\
\hline
\hline
\end{tabular}}
\end{center}
\end{table}

\paragraph{Defense against human-perceptible disturbances.}

\begin{figure}[h]
\centerline{\includegraphics[trim=0 0 0 0,clip,width=0.89\linewidth]{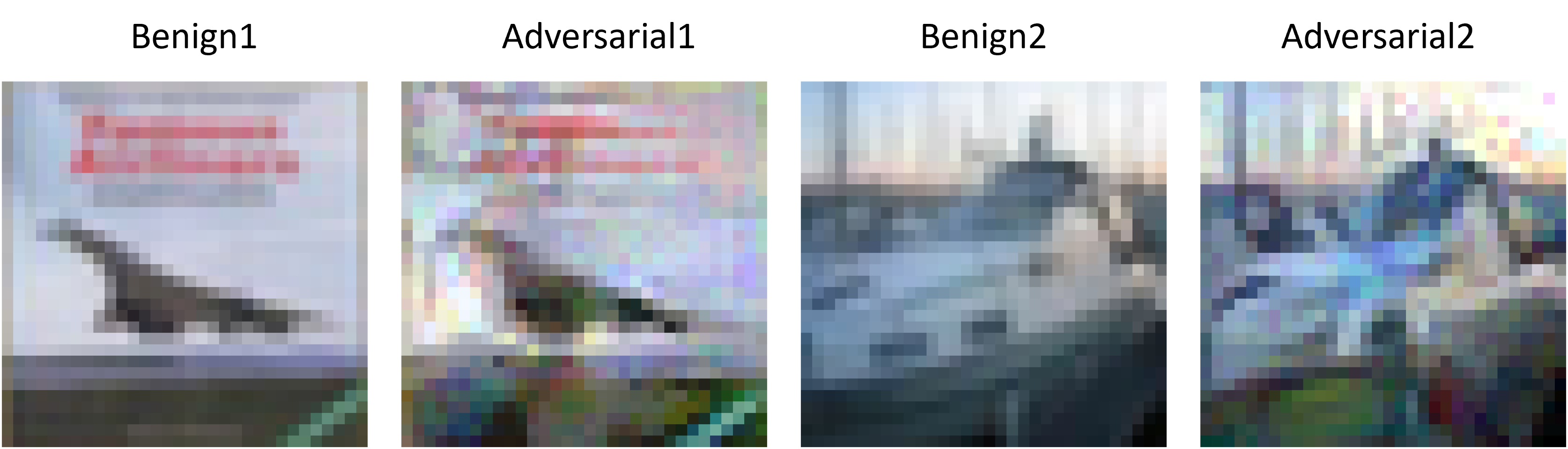}}
\caption{\textbf{Two CIFAR-10 examples of human perceptible perturbations.} The adversarial examples are generated by $\ell_\infty$-PGD attack with $\epsilon=28$.}
\label{fig: example_large_pert}
\end{figure}

We test RAILS against disturbances visible to the naked eye using CIFAR-10 data. We consider the $\ell_\infty$-PGD attack with $\epsilon=28$. Figure~\ref{fig: example_large_pert} shows the benign examples and their adversarial counterparts with large disturbances. The differences can be clearly observed. Under the human perceptible attack, the accuracy for RAILS, DkNN, and the adversarially trained CNN are $33.26\%$, $19.53\%$, and $0\%$. The results demonstrate that RAILS can effectively defend against human perceptible perturbations as compared with DkNN and CNN.

\paragraph{Diversity versus specificity.}
The DkNN finds a group of feature space k-nearest neighbors that at each layer classify 
an input sample  in a single shot. 
In contrast, starting from a initial uniform label distribution at each layer, RAILS constructs a classifier after maturation of several generations of feature representing B-cells  
using an evolutionary optimization process. Results in Table~\ref{tab2: acc_overall} show that evolving a population of features from highly diverse to highly specific provides additional robustness with little sacrifice on benign accuracy.

\paragraph{Ablation study.}
Using the CIFAR-10 dataset and the third convolutional layer of a VGG16 model, we perform an ablation study to clarify the relative influence of different RAILS components on performance. Our findings are summarized as follows: \textbf{(i)} increasing the number of nearest neighbors in a certain range improves performance ; \textbf{(ii)} ; higher mutation probability  increases robust accuracy \textbf{(iii)} ; the magnitude of mutation is sensitive to the input data, but may be optimized to increase robust accuracy. We refer readers to Appendix~\ref{sec: app6} for more details on the ablation study. We also show that when we turn off the affinity maturation stage, the robust accuracy drops from $59.2\%$ to $55.65\%$ (on $2000$ test examples), indicating the importance of including the affinity maturation step in RAILS.

\paragraph{Single-Stage Adaptive Learning.}

In the previous sections we demonstrated  that static learning is effective in predicting the class of current adversarial inputs. Here we show that RAILS can be implemented with single-stage adaptive learning (SSAL) to further improve accuracy and robustness. While the idea is not pursued in this paper, our SSAL results suggest that RAILS may be gainfully extended to the on-line learning setting. 
%
SSAL is implemented as follows. We first train a RAILS classifier on the training data as described in previous sections. Then we used  RAILS to generate $3000$ memory data (B-cells) when a subset of test data taken from MNIST was used as input to the initially trained RAILS. We then merged this new data with the population of training data, creating an augmented training set. Finally, we randomly selected and adversarially modified another $1000$ test samples of MNIST, and, using RAILS with its expanded training data, evaluated its adversarial classification accuracy. Table~\ref{tab4: acc_cnn_hard} shows that the SSAL improves RA of DkNN by $2.3\%$ with no SA loss using by augmenting the training data with only $3000$ memory data samples (a total of $5\%$ increase of the training data).

\begin{table}[h]
\begin{center}
\caption{When implemented with memory data and single-stage adaptive learning (SSAL), RAILS hardens DkNN against future attacks.  
}
\label{tab4: acc_cnn_hard}
\vspace{.05in}
\resizebox{0.45\textwidth}{!}{
\begin{tabular}{l||c|c}
\hline
\hline 
 & Standard Accuracy (SA) & Robust Accuracy (RA) ($\epsilon=60$) \\
\hline
DkNN & 98.5\%  & 68.3\%  \\
\hline
\bf{DkNN-SSAL} & 98.5\%  & \bf{70.6\%}  \\

\hline
\hline
\end{tabular}}
\end{center}
\end{table}

\section{Conclusion}
Inspired by the immune system, we proposed a new defense framework for deep learning models. The proposed Robust Adversarial Immune-inspired Learning System (RAILS) has a one-to-one mapping to a simplified architecture immune system and its learning behavior aligns with {\em in vitro} biological experiments. RAILS incorporates static learning and adaptive learning, contributing to a robustification of predictions and dynamic model hardening, respectively. The experimental results demonstrate the effectiveness of RAILS. We believe this work is fundamental and delivers valuable principles for designing robust deep models. In future work, we will dig deeper into the mechanisms of the immune system's adaptive learning (life-long learning) and covariate shift adjustment, which will be  consolidated into our computational framework. 

\appendices

\section{A one-to-one mapping from the immune system to RAILS}\label{sec: app1}
Table~\ref{tabS: map_imm_rails} provides a detailed comparison between the Immune System and RAILS. The top part shows the detailed explanations of some technical terms. The bottom part shows the four-step process of the two systems.
\begin{table*}[h]
\begin{center}
\caption{
A one-to-one mapping from the immune system to RAILS.}
\label{tabS: map_imm_rails}
\vspace{.04in}
\resizebox{0.83\textwidth}{!}{
\begin{tabular}{c||c|c}
\hline
\hline
&  Immune System &  RAILS \\
\hline
\hline
Antigen & A molecule or molecular structure (self/non-self) & Test example (benign/adversarial) \\
\hline
Affinity & \begin{tabular}[c]{@{}c@{}}The strength of a single bond or interaction\\ between antigen and B-Cell   \end{tabular}  & \begin{tabular}[c]{@{}c@{}}The negative Euclidean distance between\\ feature maps of input and another data point   \end{tabular}  \\
\hline
Naive B-cells & \begin{tabular}[c]{@{}c@{}}The B-cells that have been recruited \\ to generate new B-cells   \end{tabular}  & \begin{tabular}[c]{@{}c@{}}The k-nearest neighbors from each class \\ with highest affinity to the antigen   \end{tabular}  \\
\hline
Plasma B-cells & \begin{tabular}[c]{@{}c@{}} Newly generated B-cells with \\ top affinity to the antigen   \end{tabular}  & \begin{tabular}[c]{@{}c@{}} Newly generated examples with  \\ top affinity to the input  \end{tabular}  \\
\hline
Memory B-cells & \begin{tabular}[c]{@{}c@{}}Generated B-cells with \\ moderate-affinity to the antigen   \end{tabular}  & \begin{tabular}[c]{@{}c@{}} Generated examples with  \\ moderate-affinity to the input  \end{tabular}  \\
\hline
\hline
&  Immune System &  RAILS \\
\hline
\hline
Sensing & Classify between \textbf{self} and \textbf{non-self} antigens & \begin{tabular}[c]{@{}c@{}}Classify between \textbf{non-adversarial} and \textbf{adversarial} inputs\\ using confidence scores   \end{tabular}  \\
\hline
Flocking & \begin{tabular}[c]{@{}c@{}}Non-self antigens are presented to T cells,\\ recruit \textbf{highest affinity naive B-cells}  \end{tabular}   & \begin{tabular}[c]{@{}c@{}}Find the \textbf{nearest neighbors from each class}\\ that have the highest initial affinity score to the input data  \end{tabular}  \\
\hline
Affinity maturation & \begin{tabular}[c]{@{}c@{}}\textbf{Naive B-cells} \textit{divide and mutate} \\ to generate initial diversity. \\ \textbf{Affinity is maximized} through selection\\ by T cells for affinity.  \end{tabular}  &  \begin{tabular}[c]{@{}c@{}}Generate new examples from the \textbf{nearest neighbors} \\ through \textit{mutation and crossover} \\ and calculate each example's affinity score to the input \\ \textbf{Affinity is maximized} through selection. \end{tabular}  \\
\hline
Consensus & \begin{tabular}[c]{@{}c@{}} \textit{Memory B-cells} are saved\\ and \textit{Plasma B-cells} are created. \\ Antigen is recognized by \textbf{majority voting},\\ producing \textit{high affinity B-cells}   \end{tabular}   & \begin{tabular}[c]{@{}c@{}} Select generated examples with\\ \textbf{high-affinity scores} to be \textit{Plasma data},\\ and examples with moderate-affinity scores\\ saved as \textit{Memory data}. \\ \textit{Plasma data} use \textbf{majority voting for prediction}   \end{tabular}  \\
\hline
\hline
\end{tabular}}
\end{center}
\end{table*}


\section{Experimental parameter settings}\label{sec: app2}

\subsection{Datasets and models}
We test RAILS on three public datasets: MNIST \cite{lecun1998gradient}, SVHN \cite{netzer2011reading}, CIFAR-10 and CIFAR-100 \cite{Krizhevsky2009learning}. The MNIST dataset is a $10$-class handwritten digit database consisting of $60000$ training examples and $10000$ test examples. The SVHN dataset is another benchmark that is obtained from house numbers in Google Street View images. It contains $10$ classes of digits with $73257$ digits for training and $26032$ digits for testing. CIFAR-10 (CIFAR-100) is a more complicated dataset that consists of $60000$ ($60000$) colour images in $10$ ($100$) classes. There are $50000$ training images and $10000$ test images. We use a four-convolutional-layer neural network for MNIST, and VGG16 \cite{SZ2014} for SVHN, CIFAR-10, and CIFAR-100. For MNIST and SVHN, we conduct the affinity maturation in the inputs. Compared with MNIST and SVHN, features in the input images of CIFAR-10 and CIFAR-100 are more mixed and disordered. To reach a better performance using RAILS, we use adversarially trained models on $\epsilon=4$ and conduct the affinity maturation in layer one instead of the input layer. We find that both ways can provide better feature representations for CIFAR-10 and CIFAR-100, and thus improve RAILS performance.

\subsection{Threat models}
Though out this paper, we consider eight different types of attacks: (1) $\ell_\infty$-Projected Gradient Descent (PGD) attack \cite{madry17} - We implement $20$-step PGD attack for MNIST, and $10$-step PGD attack for SVHN and CIFAR-10. The attack strength is $\epsilon = 40/60/76.5$ for MNIST, $\epsilon = 8$ for SVHN, and $\epsilon = 8/16$ for CIFAR-10. (2) $\ell_2$-PGD attack - The attack strength is $\epsilon = 127.5$ for CIFAR-10 (3) Fast Gradient Sign Method \cite{goodfellow2014explaining} - The attack strength is $\epsilon = 76.5$ for MNIST, and $\epsilon = 4/8$ for SVHN and CIFAR-10. (4) Square Attack \cite{andriushchenko2020square} - We implement $50$-step attack for MNIST, and $30$-step attack for CIFAR-10. The attack strength is $\epsilon = 76.5$ for MNIST, and $\epsilon = 20/24$ for CIFAR-10. (5) Boundary Attack \cite{brendel2017decision} - We set the maximum iteration to be $500$ for CIFAR-10 experiments. (6) AutoAttack \cite{croce2020reliable} - The attack strength is $\epsilon=8$ and the maximum iteration is set to be $200$ for CIFAR-10 experiments. (7) Adversarial Patch \cite{brown2017adversarial} - We use patch ratio$=0.1$. (8) ASK-Attack \cite{wang2021ask} -  We implement $20$-step attack for CIFAR-10.

\subsection{Parameter selection}
By default, we set the size of the population $T=100$ and the mutation probability $\rho=0.15$. In Figure~\ref{fig: curves_running}, we set $T=100$ to obtain a better visualization. The maximum number of generations is set to $G=50$ for MNIST, and $G=10$ for CIFAR-10 and SVHN. When the model is large, selecting all the layers would slow down the algorithm. We use all four layers for MNIST. For CIFAR-10 and SVHN, we test on a few (20) validation examples and evaluate the kNN standard accuracy (SA) and robust accuracy (RA) on each layer. We then select layer three and layer four with SA and RA over $40\%$.

\paragraph{Mutation range.} The mutation range selection is related to the dataset. For MINST whose features are well separated in the input, the upper bound of the mutation range could be set to a relatively large value. For the datasets with low-resolution and sensitive to small perturbations, we should set a small upper bound of the mutation range. We also expect that the mutation could bring enough diversity in the process. Therefore, we will pick a lower bound of the mutation range. We set the mutation range parameters to $\delta_{\text{min}}=0.05 (12.75), \delta_{\text{max}}=0.15 (38.25)$ for MNIST. Considering CIFAR-10 and SVHN are more sensitive to small perturbations, we set the mutation range parameters to $\delta_{\text{min}}=0.005 (1.275), \delta_{\text{max}}=0.015 (3.825)$.

\paragraph{Sampling temperature.} Note that the initial condition found for adversarial examples could be worse than benign examples. It is still possible for examples (initial B-cells found in the flocking step) of wrong classes dominating the population affinity at the beginning. To reduce the gaps between the high-affinity examples and low-affinity examples, we use the sampling temperature $\tau$ to control the sharpness of the softmax operation. The principle of selecting $\tau$ is to make sure that the high-affinity examples in one class do not dominate the affinity of the whole population at the beginning. We thus select $\tau$ to make sure that the top $5\%$ of examples are not from the same class. We find that our method works well in a wide range of $\tau$ once the principle is reached. For MNIST, the sampling temperature $\tau$ in each layer is set to $3, 18, 18,$ and $72$. Similarly, we set $\tau=1/10$ and $\tau=300$ for the selected layers for CIFAR-10 and SVHN, respectively.

\paragraph{The hardware and our code.} We apply RAILS on one Tesla V100 with 64GB memory and 2 cores. The code is written in PyTorch.

\section{RAILS emulation of the natural immune system}\label{sec: app3}

\subsection{RAILS mimics the biological learning curve}
To demonstrate that the proposed RAILS computational system captures important properties of the immune system, we compare the learning curves of the two systems in Figure~\ref{fig: curves_bio_rails}. In RAILS, we treat test data (potentially the adversarial example) as an antigen. Affinity in both systems measures the similarity between the antigen sequence and a potential matching sequence. The green and red curves depict the evolution of the mean affinity between the B-cell population and the antigen. The candidates selected in the flocking step (kNN) are close to a particular antigen. Two tests are performed to illustrate the learning curves when the same antigen (antigen 1) or a very different antigen (antigen 2) is presented during the affinity maturation step. When antigen 1 is presented  (green curves), Figure~\ref{fig: curves_bio_rails} shows that both the immune system (left panel) and RAILS have learning curves that initially increase, then decrease, and then increase again.  This phenomenon indicates a two-phase learning process, and both systems can escape from local optimal points. On the other hand, when the different antigen 2 is presented, the flocking candidates  converge more slowly to a high affinity population during the affinity maturation process (dashed curves).

\subsection{In-vitro immune response experiment}

\begin{figure*}[h]
\centerline{\includegraphics[width=.78\textwidth]{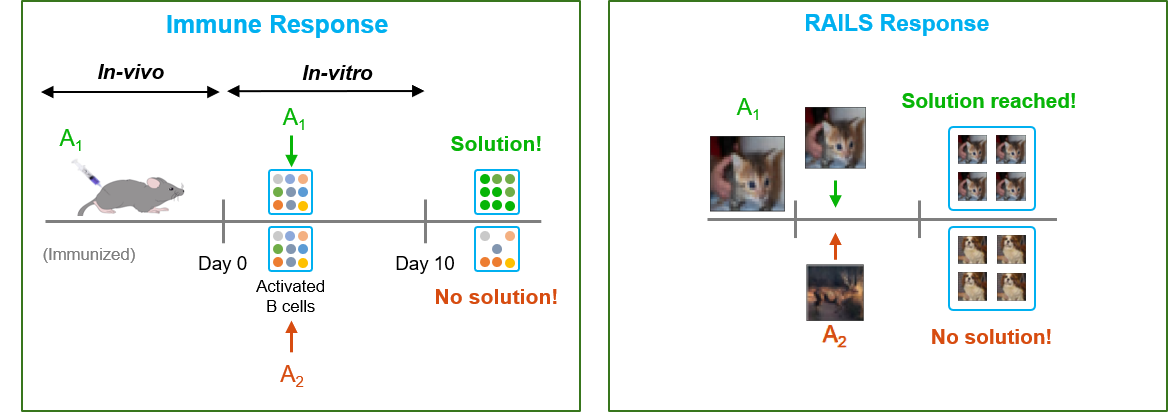}}
\caption{The conceptual diagram of in-vitro immune response (Left) and RAILS response (Right): The candidates are selected in the flocking step of antigen 1 (A$_1$). Two tests are performed to illustrate the different responses when the same antigen (A$_1$) or a very different antigen (A$_2$) is presented during the affinity maturation step. In RAILS, we treat test data (cat and deer) as antigens. The solution can be reached when A$_1$ is presented. There is no solution when A$_2$ is presented.}
\label{fig: bio_comp_exp}
\end{figure*}

\begin{figure*}[h]
\centerline{\includegraphics[width=.84\textwidth]{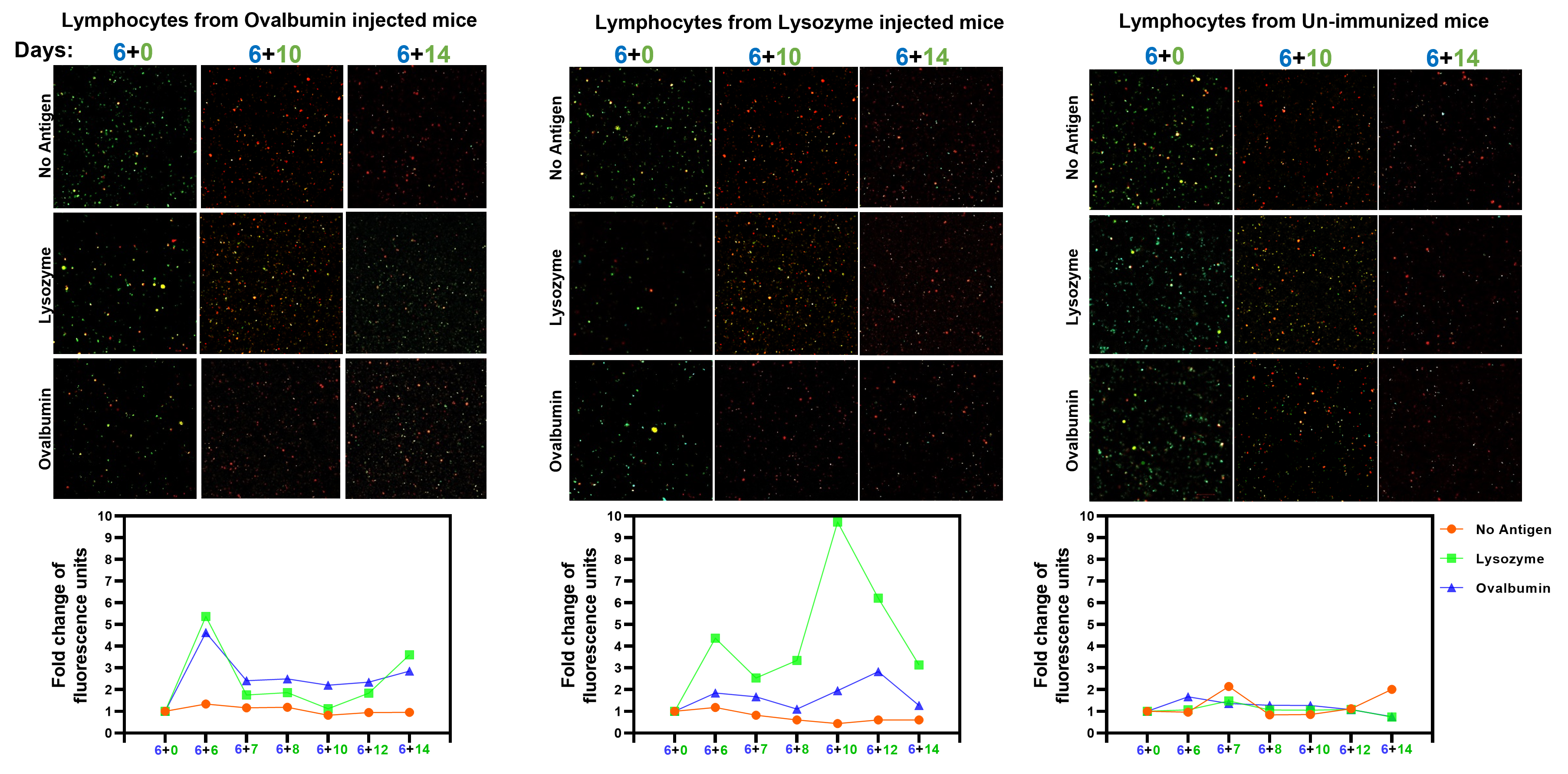}}
\caption{In-vitro experimental results showing importance of antigen affinity in robustness of adaptive immune response. The figure shows images of the B-cell proliferation (top) and B-cell fold-change curves (bottom) over time for three initial antigen imunization types (Ovalbumin-injected mice, Lysozyme injected mice, and un-immunized) and three  antigen post-imunization treatments (no-antigen, lysozyme antigen, Ovalbumin antigen) applied 6 days later.  Ovalbumin and Lysozyme antigens produce similar adaptive immune responses for those mice previously immunized with either antigen while they produce no adaptive immune response for mice that have not been immunized. The  images show clearly that proliferation of B-cells in the adaptive immune response is strongest when the lymphocytes are re-exposed to the same antigen as in the immunization but still elicits an adaptive response when exposed to a similar but non-identical antigen. The decrease in adaptive response is inversely proportional to the similarity (affinity) between the antigens.}
\label{fig: confocal_image}
\end{figure*}

\paragraph{Details on in-vitro experiments in Figure~\ref{fig: curves_bio_rails}.} We performed in-vitro experiments to evaluate the adaptive immune responses of mice to foreign antigens. These mice are engineered in a way which allows us to image their B cells during affinity maturation in an in-vitro culture. Using fluorescence of B cells, we can determine whether the adaptive immune system is effectively responding to an antigen, and infer the affinity of B cells to the antigen. In this experiment, we first immunized a mouse using lysozyme (Antigen 1). We then challenged the immune system in two ways: (1) reintroducing lysozyme and (2) introducing another very different antigen, ovalbumin (Antigen 2). We then measured the fluorescence of B cells in an in-vitro culture for each of these antigens, which we use as a proxy to estimate affinity. We use five fluorescence measurements over ten days to generate the affinity curves in Figure~\ref{fig: curves_bio_rails} (left). When plotting, we use a spline interpolation in MATLAB to smooth the affinity curves. For full experimental details, please refer to Figures~\ref{fig: bio_comp_exp}, \ref{fig: confocal_image}, and the following three sections.

\textbf{In-vitro culture of Brainbow B cells.}
For the in vitro culture of B cells, splenic lymphocytes from Rosa26Confetti+/+; AicdaCreERT2+/- mice were harvested and cultured following protocol from \cite{wand2011cooperation}. Mice were individually immunized with lysozyme and ovalbumin. Three days post-immunization, the mice were orally administered with Tamoxifen (50 µl of 20mg/ml in corn oil) and left for three days to activate the Cre-induced expression of confetti colors in germinal center B cells. Six days post-immunization, whole lymphocytes from spleen were isolated. 3$\times$105 whole lymphocytes from spleen were seeded to a single well in a 96 well dish along with 3$\times$104 dendritic cells derived from bone marrow hematopoietic stem cells for in vitro culture. The co-culture was grown in RPMI medium containing methyl cellulose (R\&D systems, MN) supplemented with recombinant IL-4 (10 ng/ml) from, LPS (1 µg/ml), 50 µM 2-mercaptoethanol, 15\% heat inactivated fetal calf serum, ovalbumin (10 µg/ml) for ovalbumin specific B cells and hen egg white lysozyme (10 µg/ml) for lysozyme specific B cells. Antigens were also added vice-versa for non-specific antigen control. The media was changed every two days. The cultures were imaged every day for 14 days.

\textbf{Preparation of differentiated dendritic cells from bone marrow hematopoietic stem cells.}
Bone marrow cells from femurs and tibiae of C57BL/6 mice was harvested, washed and suspended in RPMI media containing GM-CSF (20ng/ml), (R\&D Systems, MN), 2mM L-glutamine, 50 µM 2-mercaptoethanol and 10\% heat inactivated fetal calf serum. On day two and four after preparation, 2 mL fresh complete medium with (20ng/ml) GM-CSF were added to the cells. The differentiation of hematopoietic stem cells into immature dendritic cells was completed at day seven.

\textbf{Cell imaging.}
Confocal images shown in Figure~\ref{fig: confocal_image} were acquired using a Zeiss LSM 710. The Brainbow 3.1 fluorescence was collected at 463-500 nm in Channel 1 for ECFP (excited by 458 laser), 416-727 nm in Channel 2 for EGFP and EYFP (excited by 488 and 514 lasers, respectively), and 599-753 nm in Channel 3 for mRFP (excited by 594 laser). Images were obtained with 20× magnification.

\subsection{In-silico RAILS experiment}

\paragraph{Details on RAILS experiments in Figure~\ref{fig: curves_bio_rails}.} Similar to the in-vitro experiments, we test RAILS on two different inputs from CIFAR-10, as shown in the right panel of the conceptual diagram Figure~\ref{fig: bio_comp_exp} (cat as Antigen 1 and deer as Antigen 2). The candidates are all selected in the flocking step of antigen 1 (A$_1$). Two tests are performed to illustrate the different responses when the same antigen (A$_1$) or a very different antigen (A$_2$) is presented during the 
affinity maturation step.

We first apply RAILS on Antigen 1 (A$_1$) and obtain the average affinity of the true class as well as the initial B-cells, i.e. the nearest neighbors from all classes. The affinity vs generation curve is shown in the green line in the right panel of Figure~\ref{fig: curves_bio_rails}. One can clearly see the learning pattern. And finally, the solution is reached with a high affinity. Then we apply the initial B-cells obtained from A$_1$ to Antigen 2 (A$_2$). The results show that A$_2$ cannot reach the solution by using the given initial B-cells, as shown by the red line in the right panel of Figure~\ref{fig: curves_bio_rails}.


\section{Additional details on RAILS}\label{sec: app4}
\paragraph{Computational cost.} The RAILS prototype implemented in this paper has a relatively high computational cost, primarily due to the need to generate and select generations of in-silico B-cells using the genetic algorithm.  Specifically, the average prediction time per sample is less than $0.1$sec on CIFAR-10 with population size $100$ and $20$ generations. Note that all of our reported RAILS experiments were performed on a single GPU. RAILS speed can be dramatically accelerated by using multiple GPUs. We are currently investigating fast approximations to the genetic algorithm solution used by our prototype RAILS implementation.

\paragraph{Relations to preferential attachment} The process of selection can also be viewed as creating new nodes from existing nodes in a Preferential Attachment (PA) evolutionary graph generation process \cite{barabasi1999emergence}, where the probability of a new node linking to node $i$ is
\begin{align}\label{eq: pre_att}
    \begin{array}{ll}
\Pi(k_i) = \frac{k_i}{\sum_j k_j},
    \end{array}
\end{align}
and $k_i$ is the degree of node $i$. In PA models new nodes prefer to attach to existing nodes having high vertex degree. In RAILS, we use a surrogate for the degree, which is the exponentiated affinity measure, and the offspring are generated by parents having high degree.

\paragraph{Early stopping criterion.} Considering the fast convergence of RAILS, one practical early stopping criterion is to check if a single class occupies most of the high-affinity population for multi-generation, e.g., checking the top $5\%$ of the high-affinity population. We empirically find that it takes less than 5 iterations to convergence for most of the inputs from MNIST (CIFAR-10).

\paragraph{RAILS improves robustness of all models.} RAILS is a general framework that can be applied to any model. Specifically, we remark that there is no competitive relationship between RAILS and robust training since RAILS can improve all models' robustness, even for the robust trained model. Moreover, RAILS can reach higher robustness based on a robust trained model.


\section{A Simple Sensing Strategy}\label{sec: app5}
Sensing in the immune system aims to detect the self and non-self pieces, while RAILS leverages sensing to provide initial detection of adversarial examples. Sensing can also prevent the RAILS computation from becoming overwhelmed by false positives, i.e., recognizing benign examples to adversarial examples. Once the input is detected as benign, there is no need to go through the following process, and the neural network can directly obtain the predictions. The sensing step provides the initial discrimination between adversarial and benign inputs, and we develop a simple strategy here.

The assumption we make here is that benign examples have more consistency between the features learned from a shallow layer and the DNN prediction compared with adversarial examples. This consistency can be measured by the cross-entropy between the DNN and layer-$l$ kNN predicted class probability score. Specifically, we have the cross-entropy (adversarial threat score) for each input $\mathbf x$ in the following form
\begin{align}\label{eq: sensing}
    \begin{array}{ll}
ce(\mathbf x)^l=-\sum_{c=1}^C F_c(x) \log \mathbf v_c^l
    \end{array}
\end{align}
where $F_c$ denotes the neural network prediction score of the $c$-th class. The prediction score is obtained by feeding the output of the neural network to a softmax operation. $\mathbf v_c^l$ is the $c$-th entry of the normalized kNN vector in layer-$l$, which is defined as follows
\begin{align}\label{eq: vec}
    \begin{array}{ll}
\mathbf v_c^l=\frac{r_c}{k}=\frac{|\{\hat{x}|\hat{x} \in \mathcal{Q}_l \cap \din_{c} \}|}{k}
    \end{array}
\end{align}
where $\din_{c}$ represents the training data belonging to class $c$. $\mathcal{Q}_l$ denotes the $k$-nearest neighbors of $\mathbf x$ in all classes by ranking the affinity score $A(f_{l}; \mathbf x_j, \mathbf x)$.

\begin{figure}[h]
	\begin{minipage}[t]{0.48\linewidth}
		\centering
		\includegraphics[trim=0 0 0 0,clip,width=1\linewidth]{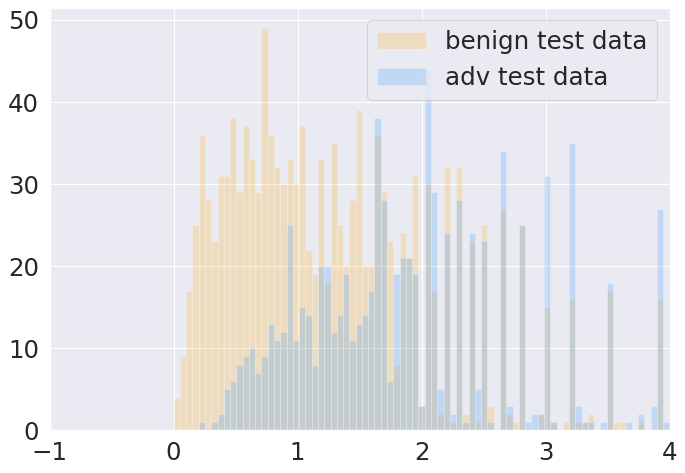}
					{\scriptsize \center (a) Distributions in layer two}
	\end{minipage}%
	\begin{minipage}[t]{0.48\linewidth}
		\centering
		\includegraphics[trim=0 0 0 0,clip,width=1\linewidth]{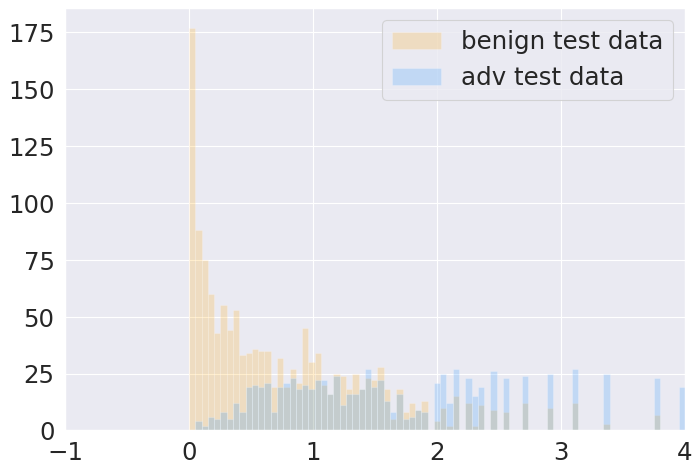}
					{\scriptsize \center (b) Distributions in layer three}
	\end{minipage}
	\caption{Adversarial threat score distributions of benign examples and adversarial examples: There are more adversarial examples with larger $ce^2$ ($ce^3$) compared to benign examples. The results suggest that a large number of benign examples could be separated from adversarial examples, and thus can prevent the RAILS computation from becoming overwhelmed by false positives.}
	\label{distl2l3}
\end{figure}

All the sensing experimental results shown below are obtained on CIFAR-10. We first compare the adversarial threat score distributions of benign examples and adversarial examples on layer two and layer three in Figure~\ref{distl2l3}. We use $1400$ benign examples for layer two and layer three. The adversarial examples are all successful attacks. One can see that there are more adversarial examples with larger $ce^2$ ($ce^3$) compared to benign examples. The results suggest that a large number of benign examples could be separated from adversarial examples, and thus can prevent the RAILS computation from becoming overwhelmed by false positives.

\begin{figure}[h]
	\begin{minipage}[t]{0.48\linewidth}
		\centering
		\includegraphics[trim=0 0 0 0,clip,width=1\linewidth]{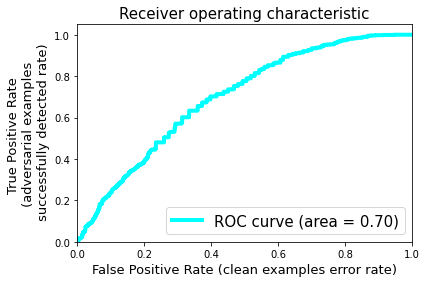}
					{\scriptsize \center (a) ROC curve of the adversarial threat scores in layer two}
	\end{minipage}%
	\begin{minipage}[t]{0.48\linewidth}
		\centering
		\includegraphics[trim=0 0 0 0,clip,width=1\linewidth]{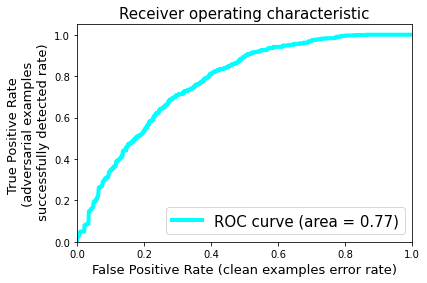}
					{\scriptsize \center (b) ROC curve of the adversarial threat scores in layer three}
	\end{minipage}
	\caption{Receiver Operating Characteristic (ROC) curves of the adversarial threat scores for benign examples and adversarial examples: The True Positive Rate (TPR) represents the adversarial examples successfully detected rate. The False Positive Rate (FPR) represents the rate of benign examples accidentally been detected as adversarial examples. The Area Under the Curve (AUC) is $0.70$ and $0.77$ for layer two and layer three.}
	\label{rocl2l3}
\end{figure} 

Figure~\ref{rocl2l3} shows the Receiver Operating Characteristic (ROC) curves of the adversarial threat scores for benign examples and adversarial examples. Figure~(a) and (b) depict the ROC curves in layer two and layer three, respectively. The True Positive Rate (TPR) represents the adversarial examples successfully detected rate. The False Positive Rate (FPR) represents the rate of benign examples accidentally been detected as adversarial examples. The Area Under the Curve (AUC) is $0.70$ and $0.77$ for layer two and layer three. Note that we care more about TPR than FPR since it has no side effect on RAILS accuracy if we detect a benign example to an adversarial example. Our goal is to select a relatively low FPR while still maintaining a high TPR. For example, we could keep a $95\%$ TPR with $56\%$ FPR using a threshold $0.4$ in layer three.

The details about the sensing algorithm are shown in Algorithm~\ref{alg:SENS_RAILS}. We will select a threshold $\kappa$ such that $x$ is treated as benign example (adversarial example) if $ce(x) \le \kappa$ ($ce(x) > \kappa$).

\begin{algorithm}[h]
   \caption{Sensing: Adversarial Example Detection}
   \label{alg:SENS_RAILS}
\begin{algorithmic}
  \STATE {\bfseries Input:} Test data point $\mathbf x$; Training dataset $\din_{tr} = \{\din_{1}, \din_{2},\cdots,\din_{C}\}$; Number of Classes $C$; Model $F$ with feature mapping $F_l(\cdot)$ in layer $l$, $l \in \mathcal{L}_s$; Affinity function $A$; A preset threshold $\kappa$ 
    \FORALL{layer $l \in \mathcal{L}_s$ \textbf{in parallel}}
  \STATE Find the $k$-nearest neighbors $\mathcal{Q}_l$ of $\mathbf x$ in all classes by ranking the affinity score $A(f_{l}; \mathbf x_j, \mathbf x)$.
  \STATE Obtain the normalized vector $\mathbf v^l = (r_1, r_2,\cdots,r_C)/k, r_c = |\{\hat{x}|\hat{x} \in \mathcal{Q}_l \cap \din_{c} \}|$.
  \STATE Obtain the softmax prediction vector ($F(x)$).
  \STATE Calculate the cross entropy $ce(\mathbf x)^l=-\sum_{c=1}^C F_c(x) \log \mathbf v_c^l$.
  \ENDFOR
  \IF {$\frac{1}{|\mathcal{L}_s|}\sum_{l \in \mathcal{L}_s}ce(\mathbf x) > \kappa$}
  \STATE $\mathbf x$ is a potential adversarial example and return $IsAdv = 1$.
  \ELSE
  \STATE $\mathbf x$ is benign and return the prediction $\arg\max_c F_c(x)$ 
  \ENDIF

\end{algorithmic}
\end{algorithm}

We then select $\mathcal{L}_s$ to only include layer three, and apply the threshold of $0.4$ in the sensing step on CIFAR-10. The results show that the false positive rate can be reduced by $40\%$, while the SA remains the same and the RA only decreases $0.2\%$.


\section{Additional Experiments}\label{sec: app6}

\subsection{Additional Comparisons on MNIST}

\begin{figure}[h]
\vspace*{-0.1in}
  \centering
  \begin{adjustbox}{max width=0.36\textwidth }
  \begin{tabular}{@{\hskip 0.00in}c  @{\hskip 0.00in}c @{\hskip 0.02in} @{\hskip 0.02in} c @{\hskip 0.02in} @{\hskip 0.02in}c }
 \begin{tabular}{@{}c@{}}  
\vspace*{0.005in}\\
\rotatebox{90}{\parbox{17em}{\centering \Large \textbf{Adv examples ($\epsilon=60$)}}}
 \\
\rotatebox{90}{\parbox{20em}{\centering \Large \textbf{Benign examples}}}
\\
\end{tabular} 
&
\begin{tabular}{@{\hskip 0.02in}c@{\hskip 0.02in}}
     \begin{tabular}{@{\hskip 0.00in}c@{\hskip 0.00in}}
     \parbox{10em}{\centering \Large Conv1}  
    \end{tabular} 
    \\
 \parbox[c]{20em}{\includegraphics[width=20em]{conv1eps60new}} 
 \\
 \parbox[c]{20em}{\includegraphics[width=20em]{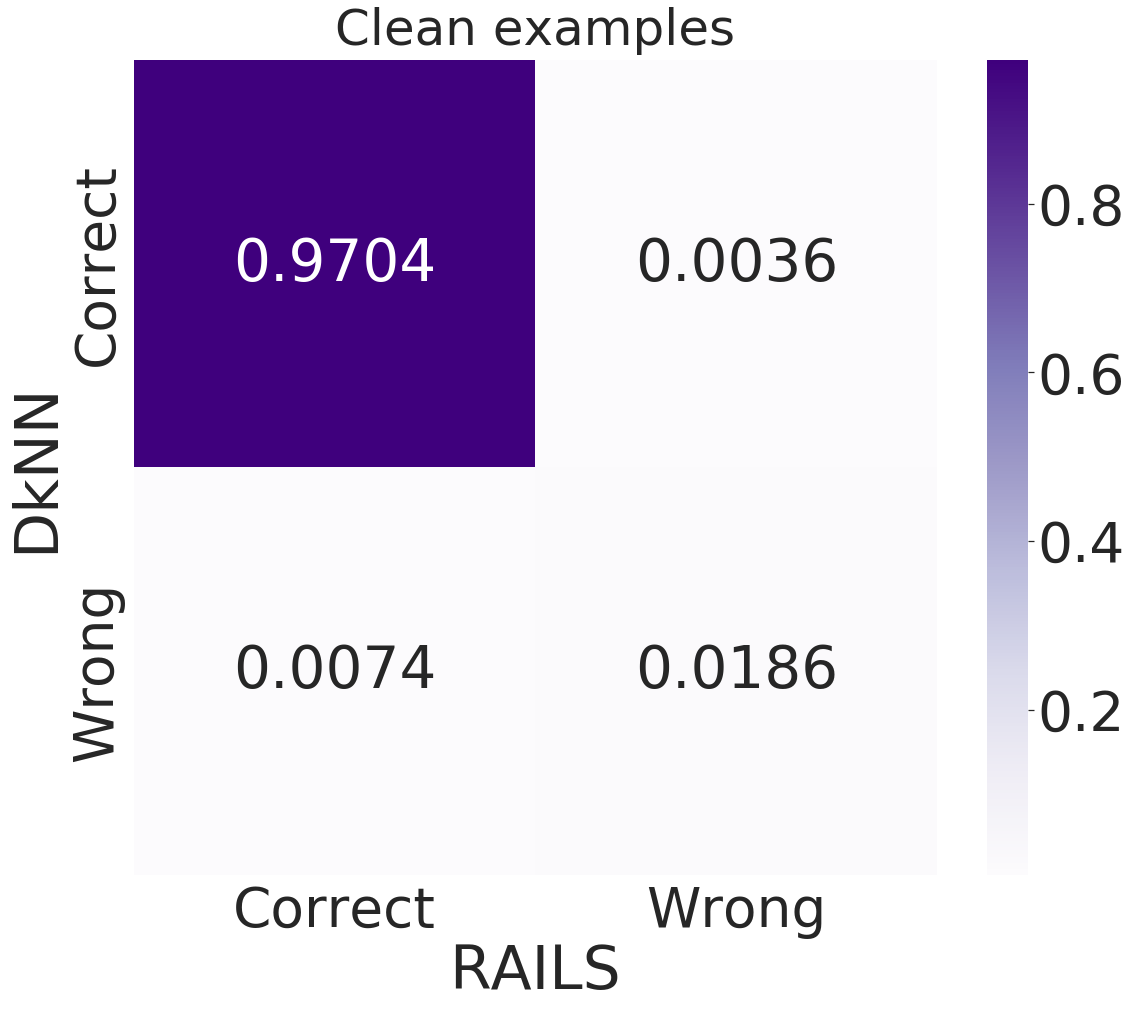}}
\end{tabular}
&
 \begin{tabular}{@{\hskip 0.02in}c@{\hskip 0.02in}}
      \begin{tabular}{@{\hskip 0.00in}c@{\hskip 0.00in}}
     \parbox{10em}{\centering \Large Conv2}
    \end{tabular} 
    \\
 \parbox[c]{20em}{\includegraphics[width=20em]{conv2eps60new}} 
 \\
 \parbox[c]{20em}{\includegraphics[width=20em]{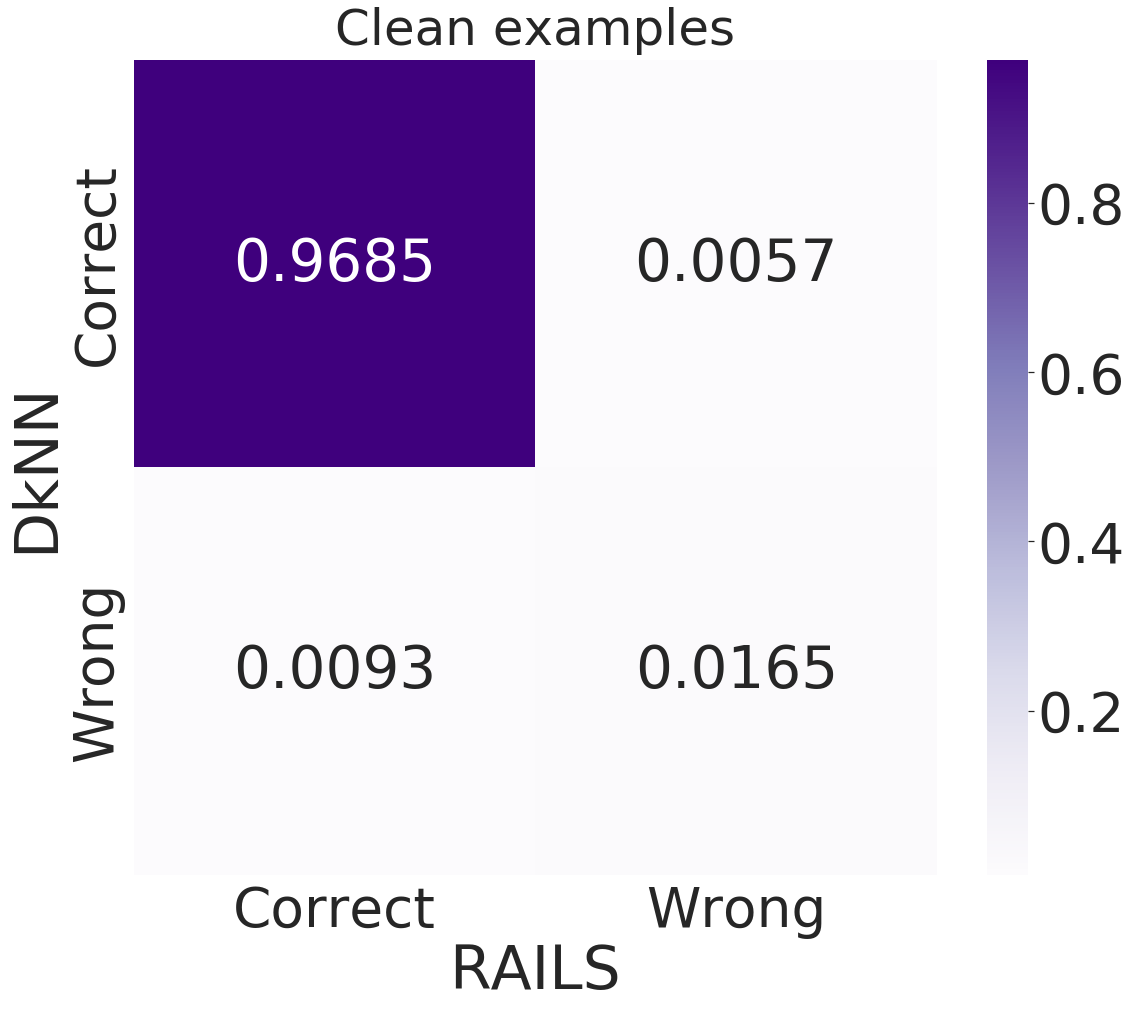}}
\end{tabular}
\end{tabular}
  \end{adjustbox}
    \caption{\textbf{RAILS has fewer incorrect predictions for those data that DkNN gets wrong.} Confusion Matrices of adversarial examples and benign examples classification in Conv1 and Conv2 (RAILS vs. DkNN $\epsilon=60$).}
  \label{figS: conf_mat}
\end{figure}

Figure~\ref{figS: conf_mat} provides the confusion matrices for benign examples classifications and adversarial examples classifications in Conv1 and Conv2 when $\epsilon=60$. The confusion matrices in Figure~\ref{figS: conf_mat} show that RAILS has fewer incorrect predictions for those data that DkNN gets wrong. Each value in Figure~\ref{figS: conf_mat} represents the percentage of intersections of RAILS (correct or wrong) and DkNN (correct or wrong).

In Table~\ref{tabS: bba}, we provide the experimental results with Square Attack  \cite{andriushchenko2020square} (a black-box attack) showing that RAILS improves the robust accuracy of DkNN by $1.35\%$ ($11\%$ attack success rate) on MNIST with $\epsilon=76.5$.

\begin{table}[h]
\begin{center}
\caption{\textbf{RAILS improves the robust accuracy of DkNN by $1.35\%$ ($11\%$ attack success rate) on MNIST with $\epsilon=76.5$.}
}
\label{tabS: bba}
\resizebox{0.32\textwidth}{!}{
\begin{tabular}{c||c|c}
\hline
\hline
&  SA&  RA\\
\hline 
\bf{RAILS (ours)}& 97.95\% & \bf{89.35\%}  \\
\hline
DkNN& 97.99\%  & 88\% \\
\hline
\hline
\end{tabular}}
\end{center}
\end{table}

\begin{table}[h]
\vspace{-0.02in}
\begin{center}
\caption{\textbf{RAILS outperforms DkNN and CNN on defending PGD attack with strength $\epsilon=8$ and $\epsilon=16$ on CIFAR-10.} The difference of robust accuracy (RA) between RAILS and DkNN increases when $\epsilon$ increases, indicating that RAILS can defend stronger attacks.}
\label{tabS: cifar}
\vspace{-.06in}
\resizebox{0.44\textwidth}{!}{
\begin{tabular}{l||c|c|c}
\hline
\hline 
 & SA & RA ($\epsilon=8$) & RA ($\epsilon=16$) \\
\hline 
\bf{RAILS (ours)} & 82\% & \bf{52.01\%} & \bf{39.04\%} \\
\hline
CNN & \bf{87.26\%}  & 32.57\% & 18.25\%\\
\hline
DkNN & 86.63\% & 41.69\% & 26.21\% \\

\hline
\hline
\end{tabular}}
\end{center}
\end{table}

\begin{table}[h]
\vspace{-0.1in}
\begin{center}
\caption{\textbf{RAILS outperforms DkNN and CNN on defending FGSM attacks with strength $\epsilon=8$ and $\epsilon=16$ on CIFAR-10.}}
\label{tabS: cifar_fgsm}
\vspace{-.06in}
\resizebox{0.46\textwidth}{!}{
\begin{tabular}{l||c|c|c}
\hline
\hline 
 & SA & RA ($\epsilon=8$) & RA ($\epsilon=16$)  \\
\hline 
\bf{RAILS (ours)} & 82\% & \bf{59.7\%} & \bf{45.26\%} \\
\hline
CNN & \bf{87.26\%} & 48.52\% & 30.55\% \\
\hline
DkNN & 86.63\% & 53.46\% & 37.19\% \\

\hline
\hline
\end{tabular}}
\end{center}
\end{table}

\begin{table}[h]
\vspace{-0.1in}
\begin{center}
\caption{\textbf{RAILS achieves higher robust accuracy (RA) than DkNN on CIFAR-10  under Square Attack \cite{andriushchenko2020square} with $\epsilon=20$ and $\epsilon=24$.}
}
\label{tab3: bba}
\vspace{.04in}
\resizebox{0.46\textwidth}{!}{
\begin{tabular}{c||c|c|c}
\hline
\hline
&  SA& RA ($\epsilon=20$) & RA ($\epsilon=24$) \\
\hline 
\bf{RAILS (ours)}& 82\% & \bf{74.5\%} & \bf{72.8\%} \\
\hline
DkNN & \bf{86.63\%} & 71.3\% & 69.5\% \\
\hline
\hline
\end{tabular}}
\end{center}
\end{table}

\begin{table}[h]
\vspace{-0.01in}
\begin{center}
\caption{\textbf{RAILS achieves higher robust accuracy (RA) than DkNN on CIFAR-10  under a (customized) ASK-Attack with $\epsilon=8$.} ASK-Attack is applied on different layers.}
\label{tab: knnattack}
\vspace{.04in}
\resizebox{0.46\textwidth}{!}{
\begin{tabular}{c||c|c|c}
\hline
\hline
&  RA (Conv 3) & RA (Conv 4) & RA (Conv 3, 4) \\
\hline 
\bf{RAILS (ours)}& \bf{41.2\%} & \bf{36.6\%} & \bf{45.5\%} \\
\hline
DkNN & 34\% & 34.4\% & 37.8\% \\
\hline
\hline
\end{tabular}}
\end{center}
\end{table}

\begin{figure}[h]
\vspace{-0.03in}
\centering
\begin{minipage}{0.46\linewidth}
\centerline{
\includegraphics[width=.86\linewidth]{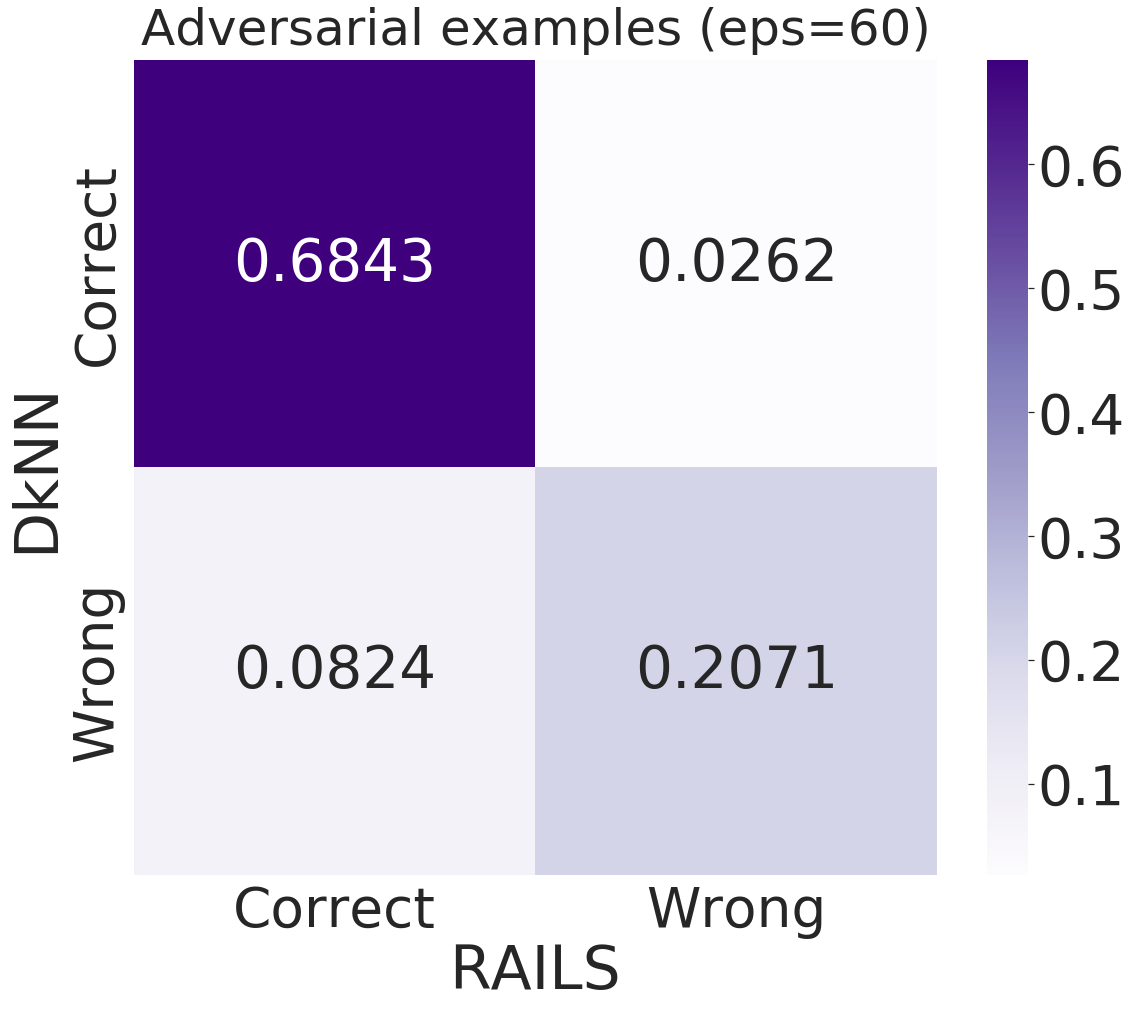}}
\end{minipage}
\centering
\begin{minipage}{0.46\linewidth}
\centerline{
\includegraphics[width=.86\linewidth]{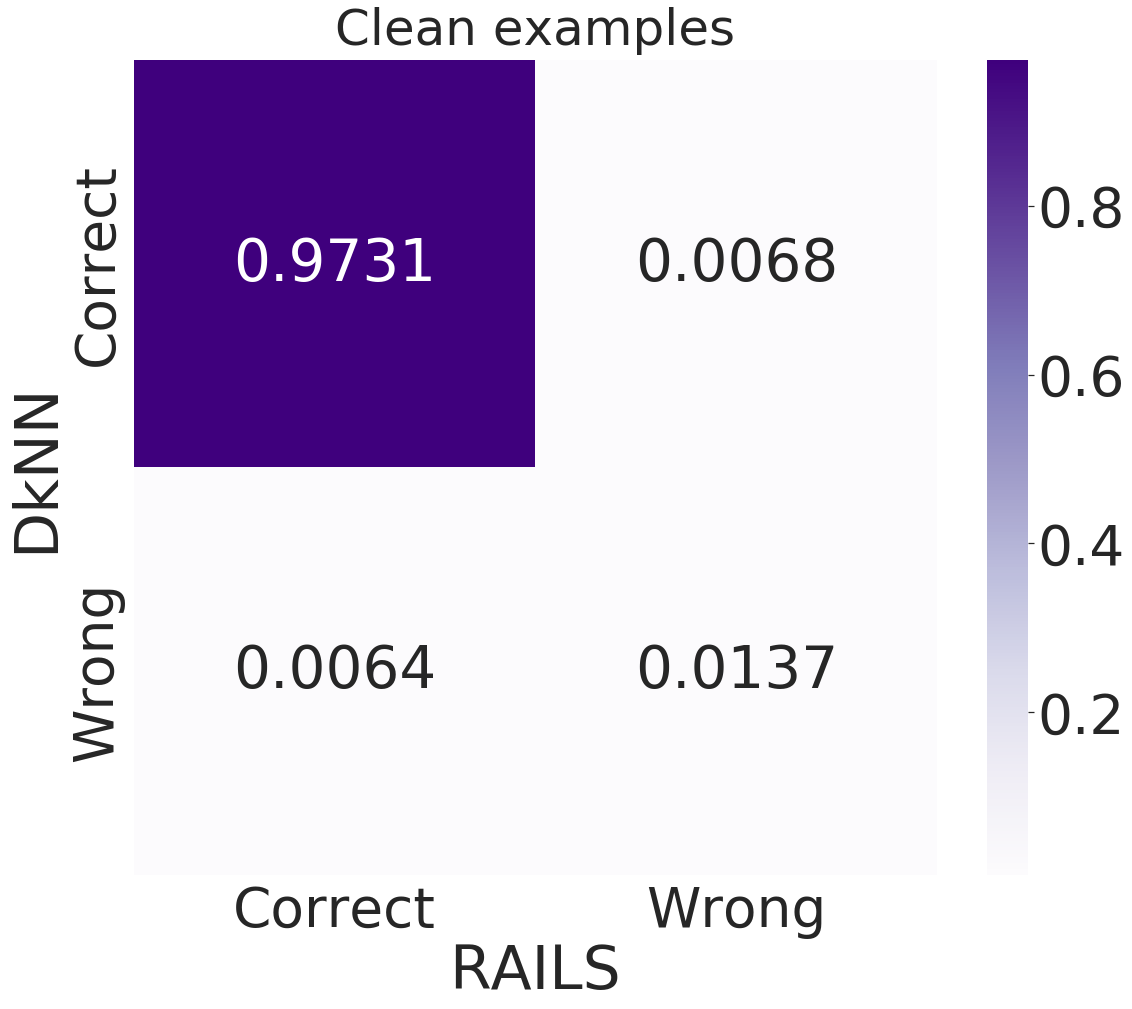}}
\end{minipage}
\caption{\textbf{RAILS has fewer incorrect predictions for those data that DkNN gets wrong.} Confusion Matrices of adversarial examples and benign examples classification (RAILS vs. DkNN - $\epsilon=60$).}
\label{fig: conf_mat_overall}
\end{figure}

Figure~\ref{fig: conf_mat_overall} shows the confusion matrices of the overall performance when $\epsilon=60$. The confusion matrices indicate that RAILS' correct predictions agree with a majority of DkNN's correct predictions and disagree with DkNN's wrong predictions.

We also show the SA/RA performance of RAILS under PGD attack and Fast Gradient Sign Method (FGSM) when $\epsilon=76.5$. The results in Table~\ref{tabS: mnist_pgd_fgsm} indicate that RAILS can reach higher RA than DkNN with close SA.

\begin{table}[h]
\begin{center}
\caption{\textbf{RAILS can reach higher robust accuracy (RA) than DkNN with similar standard accuracy (SA) on defending PGD attack and FGSM  ($\epsilon=76.5$) on MNIST.}}
\label{tabS: mnist_pgd_fgsm}
\vspace{-.04in}
\resizebox{0.46\textwidth}{!}{
\begin{tabular}{c||c|c|c}
\hline
\hline
&  SA &  RA (PGD) & RA (FGSM)\\
\hline 
\bf{RAILS (ours)}& 97.95\% & \bf{58.62\%}  & \bf{61.67\%} \\
\hline
DkNN& 97.99\%  & 47.05\% & 52.23\%\\
\hline
\hline
\end{tabular}}
\end{center}
\end{table}

\subsection{Additional Comparisons on CIFAR-10}
In this subsection, we test RAILS on CIFAR-10 under PGD attack and FGSM with attack strength $\epsilon=8/16$. The results are shown in Figure~\ref{tabS: cifar} and Figure~\ref{tabS: cifar_fgsm}. RAILS outperforms DkNN and CNN on different attack types and strengths. We also find that the difference of RA between RAILS and DkNN increases when $\epsilon$ increases, indicating that RAILS can defend stronger attacks. 

We then conduct experiments with Square Attack, which is one of the black-box attacks. The results are provided in Table~\ref{tabS: bba} and show that RAILS improves the robust accuracy of DkNN by $3\%$ on CIFAR-10 with $\epsilon=20$ and $\epsilon=24$.

\begin{table}[H]
\begin{center}
\caption{\textbf{Default parameter for RAILS.} For each parameter of interest all other parameters are set to the default values listed in this table.}
\label{tab:RAILS_default_params}
    \begin{tabular}{l||c|c}
        \hline
        \hline
        Parameter & Default Value & Layer \\
        \hline
        $C$ Classes & 10 & 3 \\
        $N$ Neighbors & 10 & 3 \\
        Population Size &  $N$ Neighbors & 3 \\
        Sampling Temperature & 1 & 3\\
        Max generation & 10 & 3 \\
        Mutation Range & (0.005 - 0.015) & 3 \\
        \hline
        \hline
    \end{tabular}
\end{center}
\end{table}

\begin{table}[H]
\begin{center}
\caption{\textbf{Increasing the $N$ neighbors within a certain range yields improved standard accuracy (SA) and robust accuracy (RA).} 
We hold the population size as a fixed $N$ neighbor value (it's minimum possible value). Note that both the standard and robust accuracy improve as the number of neighbors is increased. This may suggest that performance is influenced by the `depth` of the class-conditional selection of benign examples.}
\label{tab:n_neighbors}
    \begin{tabular}{c|c|c|c|c}
    \hline
    \hline
    SA &  RA &  Difference &  $N$ Neighbors &  Population Size \\
    \hline
     0.702 &    0.535 &     0.167 &            2 &            2 \\
     0.732 &    0.596 &     0.136 &            6 &            6 \\
     0.755 &    0.588 &     0.167 &           10 &           10 \\
     0.751 &    0.601 &     0.150 &           14 &           14 \\
     0.764 &    0.594 &     0.170 &           18 &           18 \\
     0.758 &    0.586 &     0.172 &           22 &           22 \\
     0.752 &    0.606 &     0.146 &           26 &           26 \\
    \hline
    \hline
    \end{tabular}
\end{center}
\end{table}

\begin{table}[H]
\vspace{-.05in}
\begin{center}
\caption{\textbf{Increasing population size improves robustness when $N$ is small, and does not yield significant improvement when $N$ is large with low mutation magnitude and low mutation probability.} We present an increased population coefficient $\kappa$, where the population $T$ equals $\kappa$ $\times (N$ neighbors). We present two cases: where $N$ neighbors is small and where $N$ neighbors is large. \textbf{Small $N$}: Increasing $\kappa$ from one to two improves the robustness. However, further increasing $\kappa$ does not bring significant improvement. \textbf{Large $N$}: Note that population size does not have an apparent impact on either standard or robust accuracy when $N$ neighbors is large. This may suggest that the number of perturbed input `exemplars` does not lead to more robust accuracy on adversarial inputs without sufficient mutation. This is consistent with the core principal in the adaptive immune system that mutation is necessary to converge on optimal solutions.
}
\label{tab:population_size}
    \begin{tabular}{c|c|c|c|c}
    \hline
    \hline
    SA &  RA &  Difference &  $N$ Neighbors &  Population Size \\
    \hline
     0.696 &    0.532 &     0.164 &            2 &            2 \\
     0.720 &    0.558 &     0.162 &            2 &            4 \\
     0.727 &    0.562 &     0.165 &            2 &            6 \\
     0.720 &    0.558 &     0.162 &            2 &            8 \\
     0.735 &    0.562 &     0.173 &            2 &           10 \\
     0.740 &    0.604 &     0.136 &           10 &           10 \\
     0.753 &    0.595 &     0.158 &           10 &           20 \\
     0.754 &    0.597 &     0.157 &           10 &           30 \\
     0.755 &    0.598 &     0.157 &           10 &           40 \\
     0.767 &    0.595 &     0.172 &           10 &           50 \\
    \hline
    \hline
    \end{tabular}
\end{center}
\end{table}

\begin{table}[H]
\vspace{-.05in}
\begin{center}
\caption{\textbf{Increasing mutation upper bound within a small value window increases robustness}. We observe that over several experiments mutation range has a narrow window where it positively impacts both standard accuracy (SA) and robust accuracy (RA). This may suggest that this parameter is an important hyper-parameter to tune during training.}
\label{tab:mut_range}
    \begin{tabular}{c|c|c|c|c}
    \hline
    \hline
     SA &  RA &  Difference & Mutation Range LB &  Mutation Range UB \\
    \hline
     0.764 &    0.569 &     0.195 &  0.002 &   0.005 \\
     0.740 &    0.586 &     0.154 &  0.002 &   0.010 \\
     0.746 &    0.581 &     0.165 &  0.002 &   0.015 \\
     0.743 &    0.577 &     0.166 &  0.002 &   0.020 \\
     0.732 &    0.587 &     0.145 &  0.002 &   0.025 \\
     0.737 &    0.583 &     0.154 &  0.002 &   0.030 \\
    \hline
    \hline
    \end{tabular}
\end{center}
\end{table}

\begin{table}[H]
\begin{center}
\caption{\textbf{Crossover is an important mechanism for improving performance.} We observe better performance when we use cross-over as opposed to mutation alone during training. We also note that population size alone does not necessarily contribute to better performance for either strategy.}
\label{tab:operator_type}
    \begin{tabular}{l|c|c|c|c}
    \hline
    \hline
    Operator & SA &  RA &  Difference   &  Population Size  \\
    \hline
      crossover & 0.736 &    0.570 &     0.166 &  10  \\
         mutation & 0.708 &    0.552 &     0.156 &  10  \\
      crossover & 0.744 &    0.589 &     0.155 &  20  \\
         mutation & 0.724 &    0.563 &     0.161 &  20  \\
      crossover & 0.759 &    0.598 &     0.161 &  50  \\
         mutation & 0.733 &    0.558 &     0.175 &  50  \\
    \hline
    \hline
    \end{tabular}
\end{center}
\end{table}

\subsection{Details on RAILS ablation study}

For this subsection, RAILS is trained on CIFAR-10 with VGG16 as the classifier. Results are evaluated using model classification accuracy. Accuracy is compared before and after a projected gradient descent (PGD) attack on the training data with $\epsilon = 8/255$. The baseline model performance for benign data (standard accuracy) was $87.26\%$. After the training data was adversarially attacked the VGG16 accuracy (robust accuracy) fell to $32.57\%$. By implementing RAILS, we are able to achieve an robust accuracy of $54.3\%$ using the parameterization described in Table~\ref{tab:RAILS_default_params}. All the experiments are conducted on convolutional layer $3$. Each experiment holds these parameters fixed while exploring a range of values over independent training regimes. Both standard and robust accuracy are compared for each parameter choice. The purpose of this section is to investigate RAILS' sensitivity towards parameter choices. Details for each experiment are listed in the table captions.

\bibliography{reference,refs_adv}
\bibliographystyle{IEEEtran}

\begin{IEEEbiography}[{\includegraphics[width=1in,height=1.25in,clip,keepaspectratio]{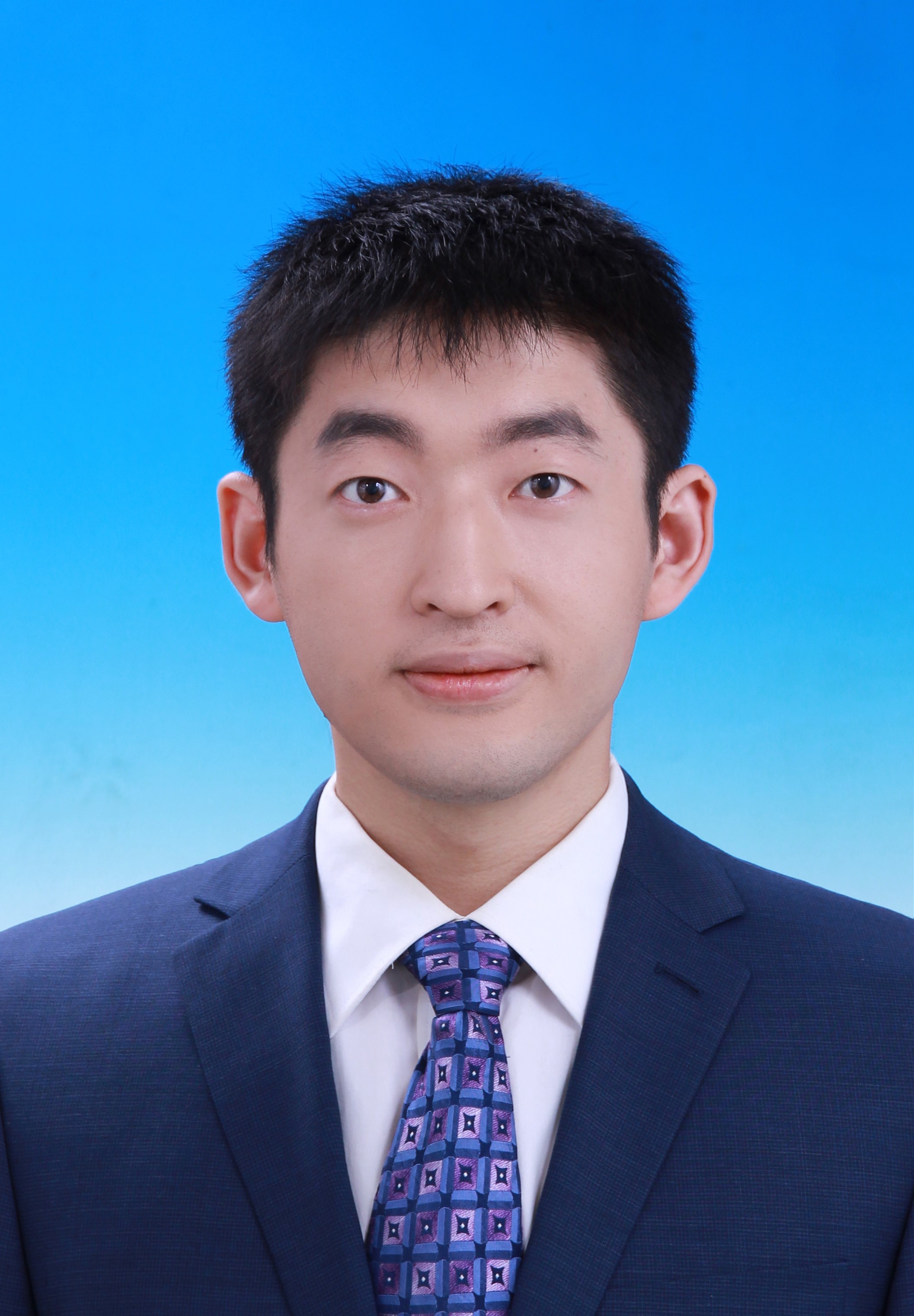}}]{Ren Wang} received bachelor’s degree and master’s degree in Electrical Engineering from Tsinghua University, Beijing, China, in 2013 and 2016. He received his Ph.D. degree in Electrical Engineering from Rensselaer Polytechnic Institute, Troy, NY, USA, in 2020. He is now a postdoctoral research fellow in the Department of Electrical Engineering and Computer Science at the University of Michigan. His research interests include Trustworthy Machine Learning, High-Dimensional Data Analysis, Bio-Inspired Machine Learning, and Robustness/Optimization on Smart Grid.
\end{IEEEbiography}

\begin{IEEEbiography}[{\includegraphics[width=1in,height=1.25in,clip,keepaspectratio]{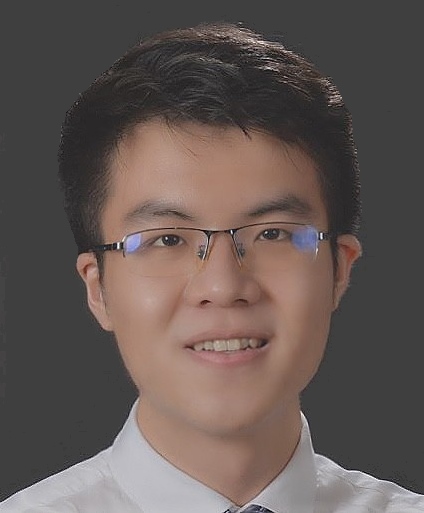}}]{Tianqi Chen} received the B.S. degree in Mathematics from Fudan University, Shanghai, China, in 2019 and the M.S. degree in Statistics from University of Michigan, Ann Arbor, MI, USA, in 2021. He is currently pursuing the Ph.D. degree in Statistics at University of Texas, Austin, TX, USA.
\end{IEEEbiography}

\begin{IEEEbiography}[{\includegraphics[width=1in,height=1.25in,clip,keepaspectratio]{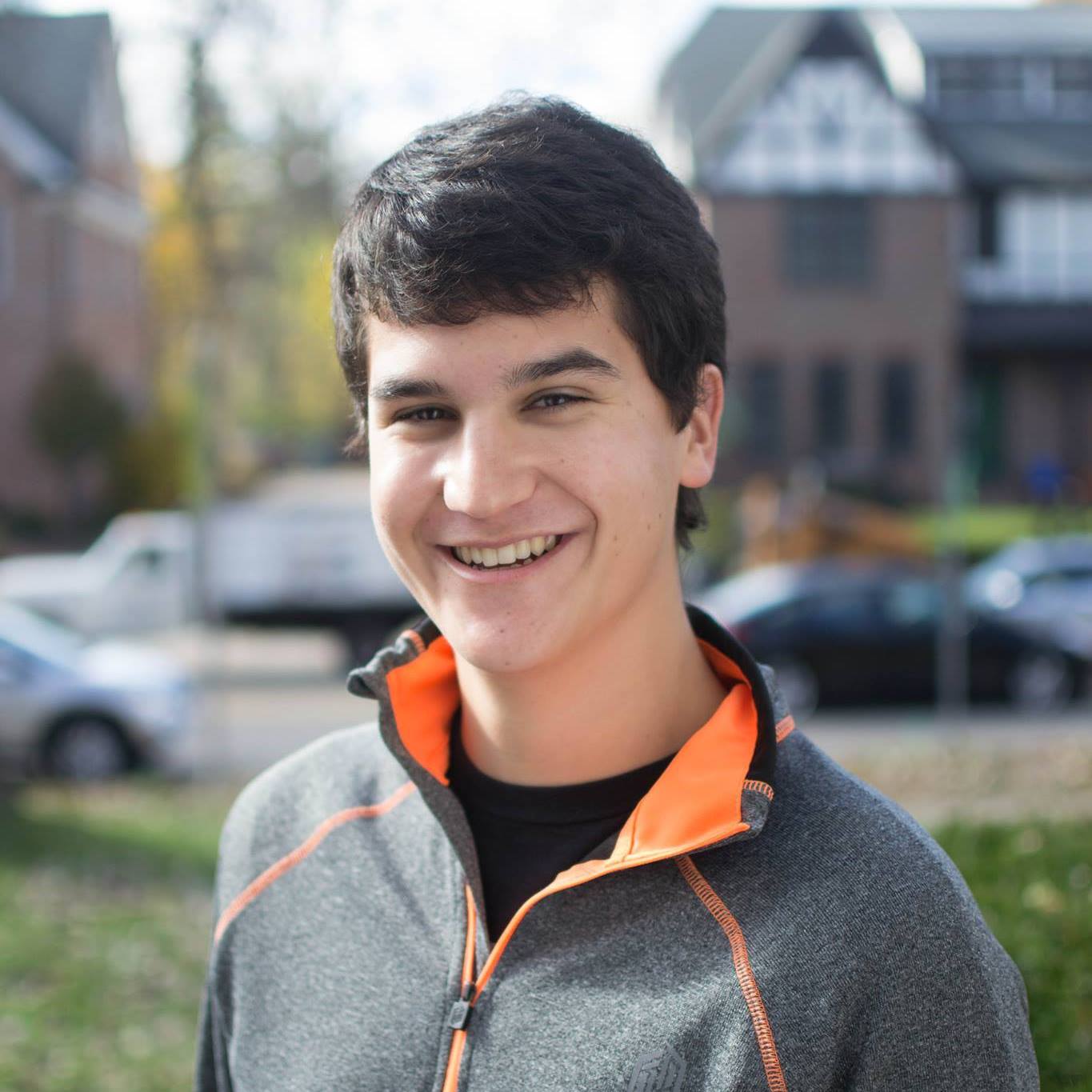}}]{Stephen M. Lindsly} was born in Midland, MI, USA in 1995. He received the B.S. degree in computer engineering in 2017 and the Ph.D. degree in bioinformatics in 2021 from the University of Michigan, Ann Arbor, MI. From 2013 to 2017, he was a Research Assistant in the Rajapakse Laboratory and a Graduate Student Instructor from 2018 to 2021. He is currently an Application Engineer at MathWorks. His research interests include the development of computational tools for the study of genome cell biology and harnessing biological systems to improve engineering design.
\end{IEEEbiography}

\begin{IEEEbiography}[{\includegraphics[width=1in,height=1.25in,clip,keepaspectratio]{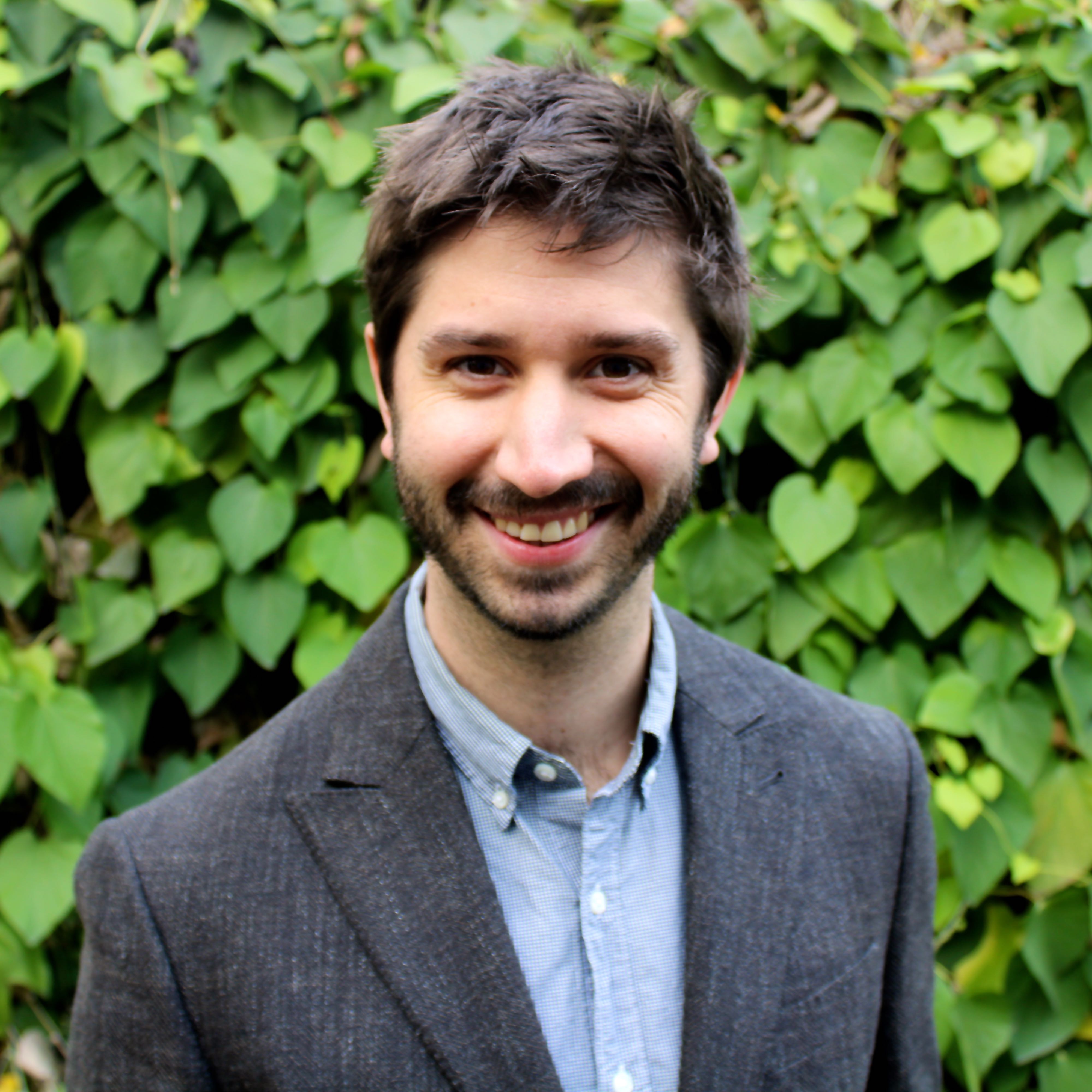}}]{Cooper M. Stansbury} was born in Ann Arbor, Michigan. He received a B.A. in Philosophy from Earlham College in 2012. He received a M.S. in Data Science from University of Michigan Dearborn in 2019 and a M.S. in Bioinformatics from University of Michigan in Ann Arbor in 2021. As of 2021 he is a PhD student in Bioinformatics in Dr. Rajapakse's Lab at the University of Michigan, Ann Arbor.
\end{IEEEbiography}

\begin{IEEEbiography}[{\includegraphics[width=1in,height=1.25in,clip,keepaspectratio]{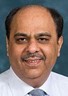}}]{Alnawaz Rehemtulla} received his MSc and PhD degrees in microbiology and immunology from the University of Calgary. Following a fellowship at the Scripps Research Institute in vascular biology, he joined the Genetics Institute in Boston as staff scientist and subsequently as faculty at the University of Michigan Medical School in 1999. He is currently Division Director for Molecular Imaging in the Department of Radiation Oncology. His research program has focused on developing and leveraging technologies for non-invasive imaging to provide novel insights into biological processes using cultured cells, mouse models and in clinical studies. 
\end{IEEEbiography}

\begin{IEEEbiography}[{\includegraphics[width=1in,height=1.25in,clip,keepaspectratio]{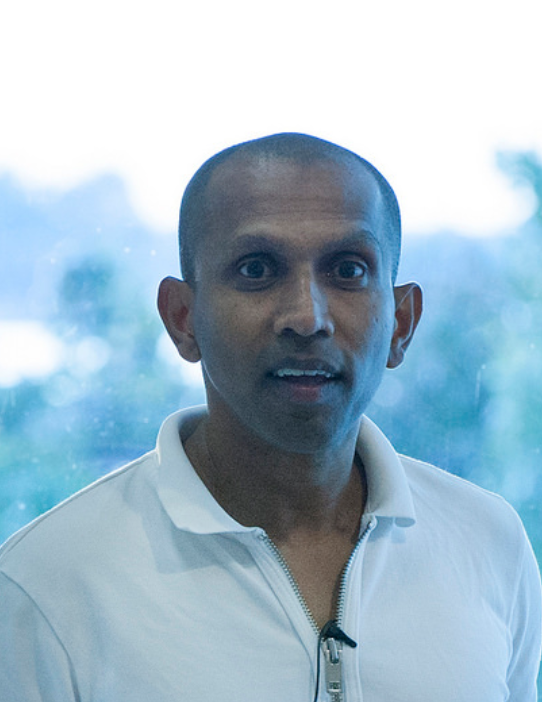}}]{Indika Rajapakse} is currently an Associate Professor of Computational Medicine $\&$ Bioinformatics, in the Medical School, and an Associate Professor of Mathematics at the University of Michigan, Ann Arbor. He is also a member of the Smale Institute. His research is at the interface of biology, engineering and mathematics. His areas include dynamical systems, networks, mathematics of data and cellular reprogramming.
\end{IEEEbiography}

\begin{IEEEbiography}[{\includegraphics[width=1in,height=1.25in,clip,keepaspectratio]{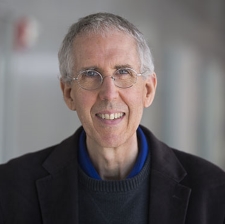}}]{Alfred O. Hero III} (F'97) received the B.S. (summa cum laude) from Boston University (1980) and the Ph.D from Princeton University (1984), both in Electrical Engineering. Since 1984 he has been with the University of Michigan, Ann Arbor, where he is the John H. Holland Distinguished University Professor of Electrical Engineering and Computer Science and the R. Jamison and Betty Williams Professor of Engineering. His primary appointment is in the Department of Electrical Engineering and Computer Science and he also has appointments, by courtesy, in the Department of Biomedical Engineering and the Department of Statistics. His recent research interests are in high dimensional spatio-temporal data, multi-modal data integration, statistical signal processing, and machine learning. Of particular interest are applications to social networks, network security and forensics, computer vision, and personalized health. He is a Section Editor of the SIAM Journal on Mathematics of Data Science and a Senior Editor of the IEEE Journal on Selected Topics in Signal Processing . He is on the editorial board of the Harvard Data Science Review (HDSR) and serves as moderator for the Electrical Engineering and Systems Science category of the arXiv. He was co-General Chair of the 2019 IEEE International Symposium on Information Theory (ISIT) and the 1995 IEEE International Conference on Acoustics, Speech and Signal Processing. He was founding Co-Director of the University’s Michigan Institute for Data Science (MIDAS) (2015-2018). From 2008-2013 he held the Digiteo Chaire d’Excellence at the Ecole Superieure d’Electricite, Gif-sur-Yvette, France. He is a Fellow of the Institute of Electrical and Electronics Engineers (IEEE) and the Society for Industrial and Applied Mathematics (SIAM). Several of his research articles have received best paper awards. He was awarded the University of Michigan Distinguished Faculty Achievement Award (2011), the Stephen S. Attwood Excellence in Engineering Award (2017), and the H. Scott Fogler Award for Professional Leadership and Service (2018). He received the IEEE Signal Processing Society Meritorious Service Award (1998), the IEEE Third Millenium Medal (2000), the IEEE Signal Processing Society Technical Achievement Award (2014), the Society Award from the IEEE Signal Processing Society (2015) and the Fourier Award from the IEEE (2020). He was President of the IEEE Signal Processing Society (2006-2008) and was on the IEEE Board of Directors (2009-2011) where he served as Director of Division IX (Signals and Applications). From 2011 to 2020 he was a member and Chair (2017-2020) of the Committee on Applied and Theoretical Statistics (CATS) of the US National Academies of Science.

\end{IEEEbiography}

\EOD

\end{document}